\pdfoutput=1
\documentclass[twoside]{article}
\usepackage[preprint]{aistats2027}

\usepackage{graphicx} %
\usepackage[utf8]{inputenc} %
\usepackage[T1]{fontenc}    %
\usepackage{hyperref}       %
\usepackage{url}            %
\usepackage{booktabs}       %
\usepackage{amsfonts}       %
\usepackage{nicefrac}       %
\usepackage{amsmath}
\usepackage{bbold}
\usepackage{subfig}
\usepackage{amssymb}
\usepackage{amsthm}
\usepackage{tikz-cd}

\usepackage{algorithm}
\usepackage{algpseudocode}
\algnewcommand\Input{\textbf{Input:}\ }
\algnewcommand\Output{\textbf{Output:}\ }
\usepackage{xpatch}
\makeatletter
\xpatchcmd{\algorithmic}
  {\ALG@tlm\z@}{\leftmargin\z@\ALG@tlm\z@}
  {}{}
\makeatother

\usepackage[
backend=biber,
style=numeric,
sorting=ynt,
url=false,
doi=false,
isbn=false,
eprint=false,
natbib=true,
uniquename=false,
maxcitenames=3,
maxbibnames=99
]{biblatex}

\defbibheading{aistats}[References]{\subsubsection*{#1}}

\newtheorem{proposition}{Proposition}

\newtheorem{lemma}[proposition]{Lemma}

\newcommand{\op}{\left(}
\newcommand{\cp}{\right)}
\newcommand{\ob}{\left[}
\newcommand{\cb}{\right]}
\newcommand{\R}{\mathbb{R}}
\newcommand{\E}{\mathbb{E}}

\newcommand{\1}{\mathbb{1}}

\newcommand{\ep}{\varepsilon}

\newcommand{\C}{\mathcal{C}}

\newcommand{\N}{\mathcal{N}}

\newcommand{\F}{\mathcal{F}}
\renewcommand{\L}{\mathcal{L}}
\newcommand{\Xb}{\mathbf{X}}
\newcommand{\M}{\mathcal{M}}
\DeclareMathOperator{\st}{stopgrad}

\newcommand{\td}{\widetilde}
\newcommand{\rv}[1]{\overleftarrow{#1}\!}
\newcommand{\fw}[1]{\overrightarrow{#1}\!}

\newcommand{\base}{\textnormal{base}}
\newcommand{\ft}{\textnormal{finetune}}
\newcommand{\target}{\textnormal{target}}
\newcommand{\noise}{\textnormal{noise}}
\DeclareMathOperator{\tr}{tr}

\DeclareMathOperator{\diag}{diag}
\DeclareMathOperator{\argmax}{argmax}

\DeclareMathOperator{\KL}{KL}

\usepackage{changes}

\begin{document}

\runningtitle{Weighted Adjoint Matching}
\runningauthor{Beyler, Bach}

\twocolumn[

\aistatstitle{Weighted Adjoint Matching: Finetuning Flow Matching Models with Time-Dependent Control Cost}

\aistatsauthor{ Eliot Beyler \And Francis Bach }

\aistatsaddress{ Inria, Ecole Normale Supérieure, PSL Research University } ]

\begin{abstract}
We introduce \emph{Weighted Adjoint Matching}, a method to finetune a pretrained diffusion or flow matching model given a differentiable reward.
We derive it from \emph{Adjoint Matching}, which reformulates finetuning as a stochastic optimal control problem with a quadratic control cost.
Building on the observation that these models generate coarse features before fine details, we place a time-dependent weight on the control cost.
A high cost early in the generative process preserves balance and diversity in the coarse features, which carry most of the semantics, while a low cost at the end lets the model increase the reward by refining details.
We prove that our method is equivalent to maximizing the reward under time-varying Kullback--Leibler regularization on the generative paths.
Experiments on Gaussian mixtures and on MNIST show a better tradeoff between reward and mode balance than Adjoint Matching.
\end{abstract}

\begin{figure*}[t]
    \centering
    \includegraphics[width=\linewidth]{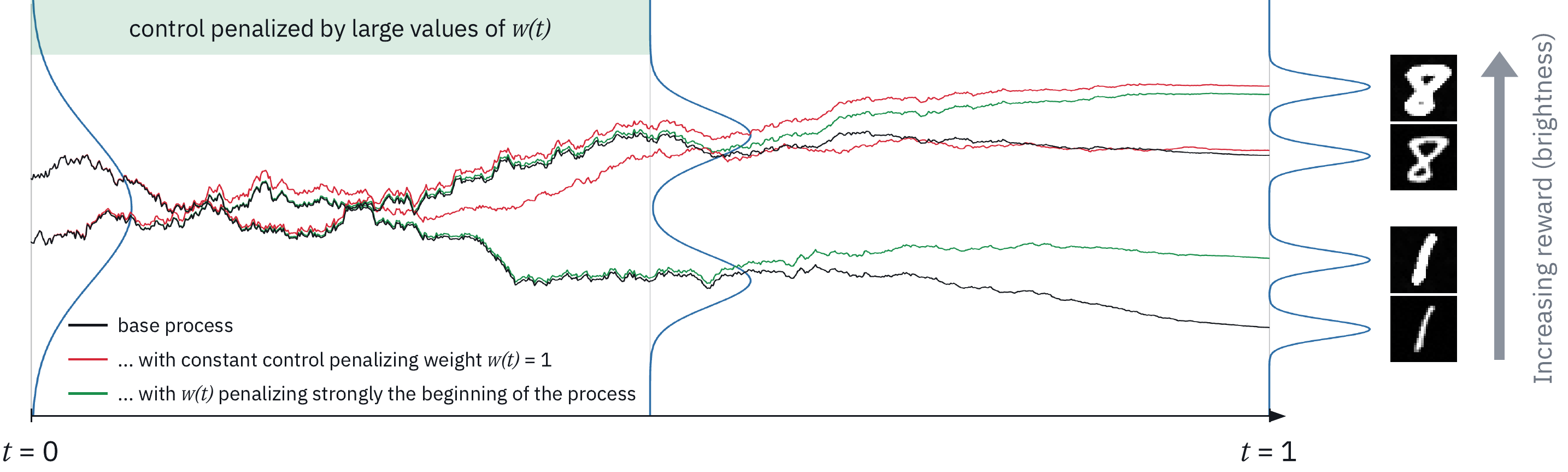}
    \caption{Trajectories of a flow matching model, in black, first sample the coarse features, illustrated here as the class of the digit, and only later its fine details, here differences in brightness. Suppose we want to finetune the model to produce brighter digits.
    Adjoint matching \citep{domingo-enrichAdjointMatchingFinetuning2025}, in \textcolor{red}{red}, penalizes the control uniformly in time.
    Early in the process, when the coarse features are not yet settled, a small push is enough to move a trajectory towards the class that is brighter on average, here the eights.
    Weighted adjoint matching, in \textcolor{green!60!black}{green}, makes the control expensive at the beginning of the process, so that the finetuned trajectories follow the base process until the coarse features are settled, then it refines only the details, leading to a more balanced distribution.}
	\label{fig:illustration_algo}
\end{figure*}

\section{Introduction}
\label{sct:introduction}
Diffusion models \citep{sohl-dicksteinDeepUnsupervisedLearning2015,songGenerativeModelingEstimating2019,hoDenoisingDiffusionProbabilistic2020,songScorebasedGenerativeModeling2021} and flow matching models \citep{lipmanFlowMatchingGenerative2023,albergoBuildingNormalizingFlows2023,liuFlowStraightFast2023} are now the state-of-the-art methods for learning and sampling a distribution known from a large dataset of empirical samples.
They work by approximating a denoising process, modeled as an ordinary or stochastic differential equation whose drift is learned with \emph{score matching} \citep{vincentConnectionScoreMatching2011}.
In this article, we focus on the task of finetuning a pretrained flow matching model, denoting $p_\target$ the distribution on $\R^d$ it samples from.

\paragraph{Finetuning a generative model.} To finetune the model, we assume that we are given a \emph{differentiable reward function} $r : \R^d \rightarrow \R$.
Our goal is to increase $r(x)$ when $x$ is a sample from the model.

A first naive way to define the finetuning could be the maximization $\max_p \E_p[r(X)]$.
But this objective makes the distribution diverge or collapse onto a maximum $x^*$ of the reward $r$%
; hence it does not capture any structure in the original data distribution $p_\target$.
Therefore, it is natural to constrain the finetuned distribution to be close to the original distribution $p_\target$, for example by penalizing the Kullback--Leibler (KL) divergence from $p_\target$: for some $a > 0$,
\begin{equation}
\label{eq:reward_objective_KL}
\max_p \E_p[r(X)] - \frac{1}{a}\KL(p\Vert p_\target).
\end{equation}
This can be shown to be equivalent to sampling from the \emph{tilted distribution} $p_a$,
\begin{equation}
\label{eq:tilted_distribution}
p_a(x) \propto p_\target(x)e^{ar(x)}.
\end{equation}

\textbf{Adjoint matching.} In practice, optimizing~\eqref{eq:reward_objective_KL} directly is intractable, and different methods aim to sample from $p_a$ by using the structure of the generative process.
The generative model can be seen as a stochastic process $dX_t = b(X_t,t)  dt + \gamma(t)dB_t$, $X_0  \sim \N(0,I)$ for some drift $b:\R^d\times \R \rightarrow \R^d$ and diffusion coefficient $\gamma: [0,1]\rightarrow \R$ such that the data distribution is the end distribution of the process, $p_1 := \L(X_1) = p_\target$.
\citet{domingo-enrichAdjointMatchingFinetuning2025} propose to reframe the finetuning as a stochastic optimal control (SOC) problem,
\begin{align}
\label{eq:soc-base}
\min_{u \in \mathcal{U}} \quad & \mathbb{E} \left[ \int_0^1 \frac{1}{2}\Vert u(X_t^u, t)\Vert^2dt - ar(X_1^u) \right] \\
\text{s.t.} \quad & dX_t^u = \left( b(X_t^u, t) + \gamma(t)\,u(X_t^u, t) \right) dt + \gamma(t)\, dB_t,\notag\\
& X_0^u \sim \N(0,I).\notag
\end{align}
This is motivated by Girsanov's theorem \citep[see, e.g.,][]{legallBrownianMotionMartingales2016}, which links the squared norm of the control and the KL divergence between the base and the controlled processes.
They develop a method to learn the optimal control $u^*$ and prove that it retrieves the tilted distribution~\eqref{eq:tilted_distribution} with $p_1^{u^*} := \L(X_1^{u^*}) = p_a$.

\paragraph{Tilting creates imbalance.} However, we may wonder if sampling from $p_a(x) \propto p_\target(x)e^{ar(x)}$ is what we want in the first place.
Indeed, this tilting has the unintended effect of moving mass between different regions of the distribution regardless of distance in sample space.
We illustrate this on a simple example\footnote{See Appendix~\ref{appendix:theory_GM} for the derivations.}
$$
\textstyle
p_\target(x) = \frac{1}{2}\N(x;-\mu,\sigma_1^2I) + \frac{1}{2}\N(x;\mu,\sigma_2^2I),
$$
with $\mu = (m,0)\in\R^2$, $m>0$ and $\sigma_2 > \sigma_1 >0$.
For $x = (x_1,x_2)$,  we define the reward $r(x) = -\frac{\Vert x_2\Vert^2}{2}$ to encourage samples to be closer to the first axis.
Note that both modes have their means $\pm\mu$ on this axis and only differ by their variances.
The tilted distribution $p_a$ is also a mixture of two Gaussian distributions
$$
\textstyle
p_a(x)= \kappa_a^1 \N(x;-\mu,\Sigma_{1,a}) + \kappa_a^2\N(x;\mu,\Sigma_{2,a})
$$
with covariance matrices $\Sigma_{i,a} = \diag\op \sigma_i^2, \frac{\sigma_i^2}{1 + a \sigma_i^2}\cp$
and weights $\kappa^1_a$ and $\kappa^2_a$, such that
$\frac{\kappa^1_a}{\kappa^2_a} = \sqrt{\frac{1 + a \sigma_2^2}{1 + a \sigma_1^2}} > 1$.
For the finetuned distribution, the first mode has a weight strictly larger than the second, creating an imbalance that was absent from the data distribution, regardless of the distance $2m$ between the centers of the modes.

\textbf{Coarse-to-fine generation in diffusion and flow matching models.} To address this issue, we rely on the fact that for natural data such as images, diffusion and flow matching models generate coarse features (the outline of an image) before fine ones (details and textures) \citep{wangDiffusionModelsGenerate2024}.
This phenomenon can be understood by looking at the frequency domain \citep{rissanenGenerativeModellingInverse2023,DiffusionSpectralAutoregression2024}: for natural images the magnitude of the (discrete) Fourier transform decays with frequency, whereas it is flat for Gaussian white noise, so progressively adding noise erases high frequencies of low magnitude before low frequencies of high magnitude.
The fact that features of high magnitude are generated before those of low magnitude can be understood within the literature on \emph{mode speciation}.
This approach to the question asks what it means for a feature to be ``generated'' at a given point in the process: past a certain time in the generative process, the feature is already determined, in the sense that almost all trajectories through the current state terminate in samples sharing it \citep{rayaSpontaneousSymmetryBreaking2023, biroliDynamicalRegimesDiffusion2024, sclocchiPhaseTransitionDiffusion2025,achilliTheorySpeciationTransitions2026}.
In particular, \citet{liCriticalWindowsNonasymptotic2024} give non-asymptotic results and analyze hierarchical data in which features live at different scales, showing that each feature emerges in a critical window of time, from features at large spatial scales to features at small spatial scales.

\textbf{Weighted adjoint matching.} Our idea is to exploit this phenomenon by modifying the SOC problem and introducing a time-dependent weight $w(t)$ on the control quadratic cost
$$
\min_{u \in \mathcal{U}} \quad  \mathbb{E} \left[ \int_0^1 \frac{w(t)}{2}\Vert u(X_t^u, t)\Vert^2dt - ar(X_1^u) \right].
$$
Modulating the control cost lets us finetune the model differently across the different spatial resolutions of the data distribution.
Typically, we penalize the early timesteps more heavily, which preserves the coarse features and lets the finetuning increase the reward by modifying the details.
The insight behind this choice is that coarse features carry most of the semantic meaning, while the reward can also be increased by tweaking only finer details.%
\footnote{While this is the general motivation behind our method, we have full flexibility to choose the timesteps that we penalize more, e.g., modify the coarse features and preserve the details, or penalize a specific frequency range.}

\textbf{Contributions.} In this paper, we make the following contributions:
\begin{itemize}
\item We introduce a new algorithm (Section~\ref{sct:weighted_adjoint_matching}), \emph{weighted adjoint matching}, to finetune flow matching and diffusion models while preserving diversity, motivated by the link between time and spatial resolutions in these models. We derive it as a simple modification of \emph{adjoint matching} \cite{domingo-enrichAdjointMatchingFinetuning2025} with no extra computational cost.
\item We show that the SOC problem defining the algorithm is equivalent to maximizing the reward under a KL regularization on every sub-path $[0,t]$ of the generative process (Section~\ref{sct:weighted_adjoint_matching:path-KL}).
\item We illustrate the method on Gaussian mixtures and on MNIST, for which the reward/mode collapse tradeoff can be precisely measured (Section~\ref{sct:experiments}).
\end{itemize}

\section{Related work}
\label{sct:related_work}
Early works aim to increase the expected value of the reward $\E_{p}[r(X)]$, often with some form of implicit regularization to limit collapse, when a model only generates very similar samples of high reward, but without characterizing the finetuned distribution.
\citet{xuImageRewardLearningEvaluating2023} introduce \emph{ImageReward}, a model trained on human preferences to score images, together with a finetuning method (ReFL) that backpropagates the reward gradient through a randomly selected step of the sampling process.
\citet{clarkDirectlyFineTuningDiffusion2024} improve upon it by back- propagating through the last $K$ steps of the process.

Several methods (see, e.g., \citep[][]{blackTrainingDiffusionModels2024,fanDPOKReinforcementLearning}) cast the finetuning of a diffusion model as reinforcement learning, viewing the denoising chain as a Markov decision process (see \citep[][]{ueharaUnderstandingReinforcementLearningBased2024} for a literature review).
However, both \citet{ueharaFineTuningContinuousTimeDiffusion2024} and \citet{domingo-enrichAdjointMatchingFinetuning2025} discuss the fact that reinforcement learning with KL regularization at each step does not in general sample from~\eqref{eq:tilted_distribution}.
\citet{ueharaFineTuningContinuousTimeDiffusion2024} take this into account by tilting the initial distribution, which requires an auxiliary model.
\citet{domingo-enrichAdjointMatchingFinetuning2025} prove that the solution of an SOC problem with quadratic control cost recovers~\eqref{eq:tilted_distribution}, provided the base process uses a particular noise schedule.
Another work by \citet{potaptchikTiltMatchingScalable2026} progressively tilts the velocity field of the base generative model to match that of the tilted distribution~\eqref{eq:tilted_distribution}.
\citet{marionImplicitDiffusionEfficient2025} propose a general framework to finetune different sampling algorithms, including unadjusted Langevin sampling and diffusion models, and in particular for the objective~\eqref{eq:reward_objective_KL}.

Some reward-finetuning methods treat the timesteps of the generative process non-uniformly as we do, but for reasons of efficiency or of gradient quality rather than to preserve diversity.
As recalled above, \citet{xuImageRewardLearningEvaluating2023} backpropagate the reward through a single, randomly selected step, chosen among the last steps of the sampling process, where they report the strongest reward signal.
\citet{clarkDirectlyFineTuningDiffusion2024} likewise restrict the backward pass to the last $K$ steps.
\citet{heTempFlowGRPOWhenTiming2026} and \citet{leePCPOProportionateCredit2026} also weight timesteps non-uniformly, but inside GRPO-style policy-gradient objectives that use only scalar rewards, and with the goal of stabilizing credit assignment across steps.
Note also that \citet{hanStochasticControlFinetuning2025} regularize each step of the sampling process with its own KL coefficient in their mathematical derivations, but only use a fixed coefficient in their experiments.

Closer to our motivation, \citet{barceloAvoidingModeCollapse2024} start from the same observation as we do, building on the works on coarse-to-fine sampling in diffusion models recalled above, and argue that treating the timesteps differently, in particular the early ones, may increase the diversity of the finetuned model by preventing mode collapse.
Similarly, after noticing that diversity mostly depends on early steps in the generative process, \citet{liuDiverseGRPOMitigatingMode2026} propose a GRPO-based finetuning method that enforces stronger regularization on those steps.
Both these methods are nonetheless based on a reinforcement-learning framework that only uses the reward as a scalar signal, and they do not characterize the finetuned distribution.
\citet{guoImprovedTechniquesFinetuning2026} build like us upon adjoint matching.
They replace the stochastic sampler by a deterministic one, which they claim to be more efficient, but therefore also lose the characterization of the finetuned distribution.
They also truncate the finetuning to the last steps of the trajectory, motivated by efficiency rather than diversity.
This differs from our approach: not computing the loss on the early timesteps does not enforce that the finetuned velocity field stays equal to the base one at those times, since a single neural network is shared across timesteps.
Our method instead ensures explicitly that the finetuned model stays close to the base model at the times where the weight $w(t)$ is large.

Finally, \citet{domingo-enrichAdjointMatchingLens2026} generalize adjoint matching to a large class of stochastic optimal control objectives, with control-dependent drift and diffusion coefficients and convex running costs.
Our time-weighted objective~\eqref{eq:soc-weighted} can be partially fitted into this framework, which recovers our weighted adjoint matching loss only for a particular choice of the stabilizing weight $\lambda(t)$.
We give more details on this connection in Appendix~\ref{appendix:weighted_adjoint_matching_algo:connection}.

\section{Generative SDEs: sampling features progressively}
\label{sct:intro_diffusion}
The goal of this section is to motivate the introduction of the time-varying weight $w(t)$ as a way to adjust the intensity of the control according to the different phases of the process.
We first introduce diffusion and flow matching models, and explain what is special about the choice of \emph{memoryless noise schedule} proposed by \citet{domingo-enrichAdjointMatchingFinetuning2025}.
We then link this choice to the mode speciation results \citep{rayaSpontaneousSymmetryBreaking2023, biroliDynamicalRegimesDiffusion2024, sclocchiPhaseTransitionDiffusion2025,liCriticalWindowsNonasymptotic2024,achilliTheorySpeciationTransitions2026} that we illustrate for a Gaussian mixture.

\textbf{Forward and reverse processes of diffusion models.}%
At the heart of diffusion and flow matching models is a Gaussian noise corruption process (see, e.g., \citep[][]{songScorebasedGenerativeModeling2021,karrasElucidatingDesignSpace2022}).
It starts from~$X\sim p_{\target}$ and adds noise to it:%
\footnote{Detailed derivations are given in Appendix~\ref{appendix:intro_diffusion}.}
\begin{equation}
\label{eq:sde_forward_gen}
\left\{
\begin{array}{l}
d\fw X_\tau =  -f(\tau)\fw X_\tau d\tau + g(\tau)dB_\tau, \\
\fw X_0 = X,
\end{array}\right.
\end{equation}
where $f,g : \R \mapsto\R_+$ are continuous functions.
For each time $\tau$, the joint distribution of $(\fw X_0,\fw X_\tau)$ is such that $\fw X_\tau = a(\tau) \fw X_0 + b(\tau) \N(0,I)$, with coefficients $a(\tau) = \exp\op-\int_0^\tau f(u)du\cp$, and $b(\tau) = a(\tau)\sqrt{\int _0^\tau \frac{g(u)^2}{a(u)^2}du}$. This joint distribution is therefore simple to study and to sample from.

A generative model is obtained by reversing this corruption process.
Choosing $T>0$ and a noise schedule $\gamma: \R \rightarrow \R^+$, and denoting $\rv f(t) = f(T-t), \rv g(t) = g(T-t), \rv p_t = \fw p_{T-t}$ the density of $\fw X_{T-t}$, and $W_t$ a Brownian motion, the process defined by 
\begin{equation}
\label{eq:general_backward_process}
\left\{
\begin{array}{@{\,}l@{}l@{}}
d\rv{X}_t &{}=  \op \rv f(t)\rv X_t + \frac{\gamma(t)^2 + \rv g(t)^2}{2} \nabla\log \rv p_t(\rv X_t)\cp dt \\
&{}+ \gamma(t) dW_t, \\
\rv{X}_0 &{}= \fw X_T.
\end{array}\right.
\end{equation}
has the same \emph{marginals} as $\fw X_\tau$: $\forall t \in [0,T], \L(\rv X_t) = \L(\fw X_{T-t})$.

Different choices of $\gamma(t)$ are possible but two are particularly remarkable. Choosing $\gamma(t) = 0$ makes the backward process an ODE, and leads to deterministic sampling methods such as DDIM \citep{songDenoisingDiffusionImplicit2021}. But the one that interests us here is $\gamma(t) = \rv g(t)$: then $\L((\rv X_t)_{t\in[0,T]}) = \L((\fw X_{T-t})_{t\in[0,T]})$ \citep{haussmannTimeReversalDiffusions1986}.
The backward process preserves not only the marginals of the forward but also the path distribution.

\textbf{Flow matching and the memoryless noise schedule.}
Following \cite{domingo-enrichAdjointMatchingFinetuning2025}, we use a flow matching model with the Gaussian source distribution.
To define it, start from $\bar X_0 \sim p_\noise =\N(0,I)$ and $\bar X_1 \sim p_\target$ two independent random variables, and for $t \in [0,1]$, $\bar X_t = \beta_t \bar X_0 + \alpha_t \bar X_1$, with $\alpha_t, \beta_t$ differentiable schedules such that $\alpha_0 = \beta_1 = 0$, $\alpha_1 = \beta_0 = 1$, $t \mapsto \alpha_t$ non-decreasing, $t \mapsto \beta_t$ non-increasing.
A flow matching model is a process
\begin{equation}
\label{eq:flow-matching-ODE}
\left\{
\begin{array}{ll}
     dX_t & = v(X_t,t) dt \\
     X_0 & \sim p_\noise = \N(0,I),
\end{array}
\right.
\end{equation}
with $v$ the velocity field defined as the minimizer of the quadratic loss $\E\ob\Vert v(\bar X_t,t) - \dot\beta_t \bar X_0 - \dot \alpha_t \bar X_1\Vert^2\cb$.
\cite{domingo-enrichAdjointMatchingFinetuning2025} transform the sampling process~\eqref{eq:flow-matching-ODE} to an SDE by adding a diffusion coefficient $\gamma(t)$
\begin{equation}
\label{eq:flow-matching-SDE}
\left\{
\begin{array}{ll}
     dX_t & = b^\gamma(X_t,t)  dt + \gamma(t)dB_t\\
     X_0 & \sim \N(0,I),
\end{array}
\right.
\end{equation}
with $b^\gamma(x,t) = v(x,t) + \frac{\gamma(t)^2}{2}\nabla\log p_t(x)$.

Though defined differently, this is just a diffusion model with a particular choice of time parametrization.
Indeed, the velocity field is linked to $\nabla\log p_t(x)$ through (see, e.g., \citep[][]{domingo-enrichAdjointMatchingFinetuning2025} Appendix~B.4)
\begin{equation}
\label{eq:relation_velocity_score} \textstyle
v(x,t) = \frac{\dot \alpha_t}{\alpha_t}x + \beta_t\op\frac{\dot\alpha_t}{\alpha_t}\beta_t - \dot\beta_t\cp \nabla\log p_t(x).
\end{equation}
Plugging this equality~\eqref{eq:relation_velocity_score} into the flow matching process~\eqref{eq:flow-matching-SDE} gives the diffusion backward process~\eqref{eq:general_backward_process} with $T=1$, and $\rv f(t),\rv g(t)$ taken such that $\alpha_t = \rv a(t) := a(1-t)$ and $\beta_t = \rv b(t) := b(1-t)$,
\begin{equation*}\textstyle
\rv f(t) = \frac{d}{dt}\log \alpha_t = \frac{\dot \alpha_t}{\alpha_t}, \quad
\rv g(t) = \sqrt{2 \beta_t \op\frac{\dot\alpha_t}{\alpha_t}\beta_t-\dot\beta_t\cp}.
\end{equation*}
In particular, choosing $\gamma(t) = 0$ recovers deterministic diffusion models, while the \emph{memoryless noise schedule},
\begin{equation}
\label{eq:memoryless_noise_schedule}
\textstyle
\gamma(t) = \sqrt{2 \beta_t \op\frac{\dot\alpha_t}{\alpha_t}\beta_t-\dot\beta_t\cp},
\end{equation}
considered by \citet{domingo-enrichAdjointMatchingFinetuning2025} is simply $\gamma(t)= \rv g(t)$, i.e., the choice for which the backward SDE preserves the joint distributions.
The memoryless property $p(X_0,X_1)=p(X_0)p(X_1)$ is just a consequence of the conservation of the joint distributions as $(X_0,X_1) = (\fw X_1, \fw X_0) =(\alpha_0 X + \beta_0 Z,X) = (Z,X)$ with $Z\sim \N(0,I)$ independent from $X$.

\begin{figure*}
    \centering
    \captionsetup[subfigure]{justification=centering}
	\subfloat[][Stochastic sampling paths.]{
    \includegraphics[width=.49\linewidth]{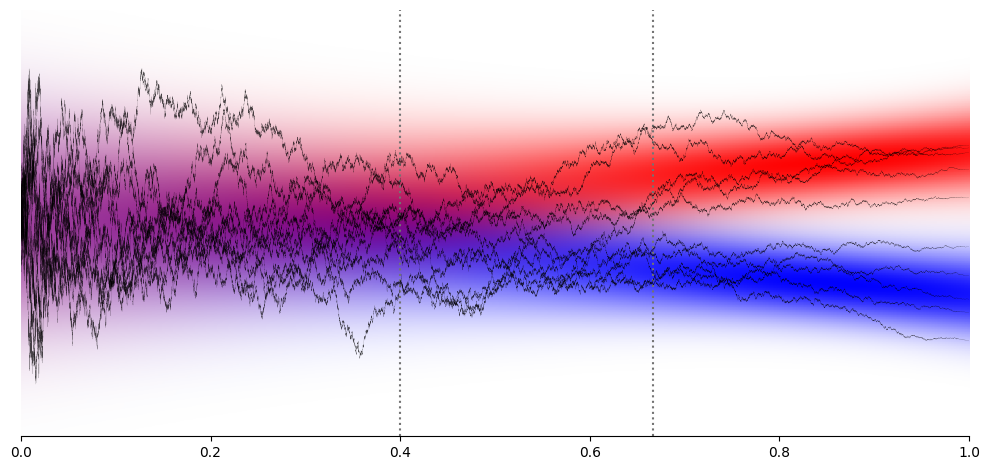}
    }
    \subfloat[][Deterministic sampling paths.]{
    \includegraphics[width=.49\linewidth]{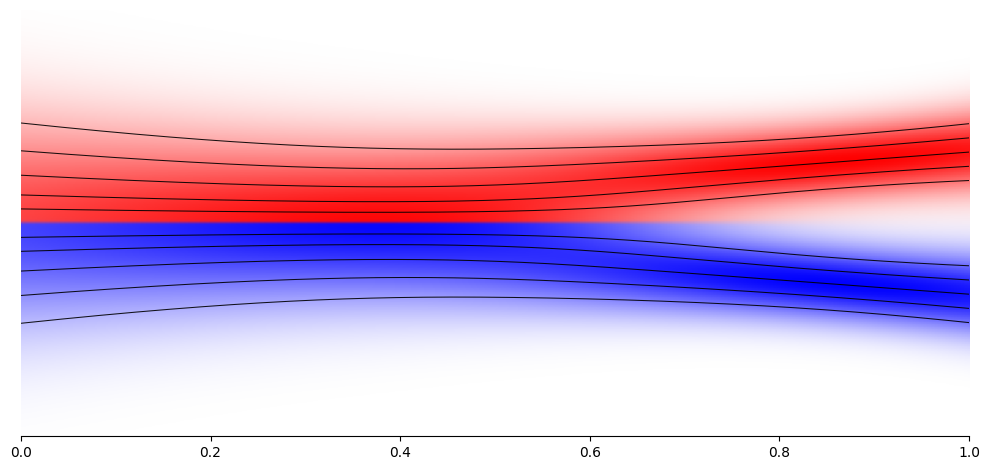}
    }
    \caption{\textcolor{red}{Red} component proportional to probability that trajectory going through the point ends up in upper mode, \textcolor{blue}{blue} to probability for lower mode, and transparency proportional to density of marginal distribution.}
	\label{fig:trajectory_mixing}
\end{figure*}

\textbf{Noising and spatial scale of the target distribution.}
In Figure~\ref{fig:trajectory_mixing}, we illustrate the advantage of the sampling paths (a) given by the memoryless noise schedule, over the original deterministic paths (b) of flow matching models.
For each point, the red component of the pixel color is proportional to the probability that a trajectory going through this point ends up in the upper mode, the blue component to the probability for the lower mode, and transparency is proportional to the density of the marginal distribution at this point.
We see that for the memoryless sampling process (a), before the time window delimited by the two gray dotted lines, the class is undetermined (mostly purple) while after this window, the two classes are distinguishable.
On the other hand, for a deterministic sampling process (b), classes are predetermined from the initial sampling at time $0$.

More generally, the times of this destroying/generating of features depends on the spatial scale of the feature \citep{rissanenGenerativeModellingInverse2023,rayaSpontaneousSymmetryBreaking2023,wangDiffusionModelsGenerate2024,DiffusionSpectralAutoregression2024,biroliDynamicalRegimesDiffusion2024,sclocchiPhaseTransitionDiffusion2025,achilliTheorySpeciationTransitions2026}.
In particular, \cite{liCriticalWindowsNonasymptotic2024} characterize these \emph{critical windows} where a feature emerges as $[T_\textrm{lower},T_\textrm{upper}]$, where both bounds depend directly on the spatial scale of the feature.
Moreover, they generalize this result for a hierarchical model, with features at different spatial scales, leading to successive time windows, starting from large spatial scales to finer ones.

\section{Weighted adjoint matching}
\label{sct:weighted_adjoint_matching}

We introduce a time-dependent weight $w:[0,1]\rightarrow \R_+^*$ in the quadratic control cost of the SOC problem~\eqref{eq:soc-base}:
\begin{align}
\label{eq:soc-weighted}
\min_{u \in \mathcal{U}} \quad & \E \left[ \int_0^1 \frac{\textcolor{red}{w(t)}}{2}\Vert u(X_t^u, t)\Vert^2 dt - a r(X_1^u) \right] \\
\text{s.t.} \quad & dX_t^u = \left( b(X_t^u, t) + \gamma(t)\,u(X_t^u, t) \right) dt + \gamma(t)\, dB_t, 
\notag
\\
& X_0^u \sim p_0,\label{eq:soc-weighted:SDE}
\end{align}
where $r$ is the reward function and $a>0$ the tilting parameter of~\eqref{eq:reward_objective_KL}.
To keep the exposition light, this section is written for the finetuning task, using a flow matching model with the memoryless noise schedule~\eqref{eq:memoryless_noise_schedule} and $b(x,t):= b^\gamma(x,t) = 2v^\base(x,t) - \frac{\dot \alpha_t}{\alpha_t}x$.
Every result stated here holds for an arbitrary running cost $f(x,t)$, terminal cost $g(x)$, and base SDE on $t\in [0,T]$ (hence including the finetuning of diffusion models), and is stated and proven in that generality in Appendix~\ref{appendix:path-KL} and Appendix~\ref{appendix:weighted_adjoint_matching_algo}.
We assume that $w$ is a function of bounded variation such that
\begin{equation}
\label{eq:weight-integral-form}
\textstyle
w(t) = w(0) + \int_0^tdw(s) = w(1) - \int_t^1dw(s).
\end{equation}
In particular, a differentiable $w$ satisfies these equalities with $dw(s) = w'(s)ds$, but the assumption also allows discontinuous weights such as piecewise constant functions.
We denote $p^u$ the probability distribution on the path $\Xb^u = (X_t^u)_{t\in [0,1]}$ defined by a control $u$ and $p^\textnormal{base} = p^{u=0}$  the one of the base generative model.
\subsection{Path-KL regularized objective}
\label{sct:weighted_adjoint_matching:path-KL}
We denote by $K_t(X_0)$ the KL divergence between the controlled and the base processes on the time interval $[0,t]$, conditionally on the initial condition,
\begin{equation}
\label{eq:path-KL}
K_t(X_0) := \KL(p^u(X_{[0,t]}|X_0)\Vert p^\base(X_{[0,t]}|X_0)).
\end{equation}
The weighted control cost of~\eqref{eq:soc-weighted} can be rewritten in terms of these path KL divergences, which turns the weighted SOC problem into an objective regularized by the KL divergence on every sub-path $[0,t]$.
\begin{proposition}
\label{prop:path-KL}
Assume that $w$ satisfies~\eqref{eq:weight-integral-form}.
Then the weighted SOC problem~\eqref{eq:soc-weighted} is equivalent to
\begin{align}
\label{eq:path-KL-regularized-objective}
&\max_{u \in \mathcal{U}} \E_{X_0 \sim p_0} \Big[ a\E_{\Xb^u\sim p^u(\cdot|X_0)} \ob r(X_1^u)
\cb \\
&\textstyle - w(1)\KL(p^u(\Xb|X_0) \Vert p^\base(\Xb|X_0)) + \int_0^1 K_t(X_0) dw(t) \Big ].\notag
\end{align}
\end{proposition}
This result is proven in Appendix~\ref{appendix:path-KL}, which also contains a dual reformulation, regularized by the KL on the sub-paths $[t,1]$, and the link between the two.

Compared to adjoint matching of \cite{domingo-enrichAdjointMatchingFinetuning2025}, the KL between the paths on $[0,t]$ enters the objective with weight $dw(t)$, which for a non-increasing $w$ is negative and therefore adds regularization. 

\textbf{Example.} To penalize more heavily the part of the process before mode speciation, we can 
take a piecewise constant function $w$, with $w_0>0$,
$$
w(t) =
\left\{
\begin{array}{ll}
     1+w_0 & \textnormal{if} \quad t< t_s   \\
     1 & \textnormal{if} \quad t >  t_s  
\end{array}
\right.
$$
for which divergence from the base distribution on $(0,t_s)$ is penalized a second time by $w_0$:
\begin{multline*}
\max_{u \in \mathcal{U}} \E_{X_0 \sim p_0} \Big[ a\E_{\Xb^u\sim p^u(\cdot|X_0)} \ob r(X_1^u)
\cb \\ -\KL(p^u(\Xb|X_0) \Vert p^\base(\Xb|X_0)) \\ - w_0 \KL(p^u(X_{[0,t_s]}|X_0)\Vert p^\base(X_{[0,t_s]}|X_0))  \Big].
\end{multline*}

\subsection{Algorithm}
\label{sct:weighted_lean_adjoint}

In this section, we derive an algorithm to learn the optimal control $u^*$ of the weighted SOC problem~\eqref{eq:soc-weighted}.
We adapt the adjoint matching method \cite{domingo-enrichAdjointMatchingFinetuning2025}.
We give here a general overview; the complete proofs of the results are given in Appendix~\ref{appendix:weighted_adjoint_matching_algo}.
We aim to learn the optimal control $u^*$ by using the fact that it satisfies
\begin{equation}
\label{eq:formula_u_grad_J}
u^* (x,t) =  -\frac{\gamma(t)}{w(t)}\nabla_x J(u^*,x,t),
\end{equation}
with the \emph{cost function} defined as
\begin{equation*}
J(u,x,t) =  \E_{\Xb \sim p^u}\ob\scriptstyle\int_t^1\frac{w(s)}{2}\Vert u(X_s,s)\Vert^2 ds - a r(X_1)\Big | X_t = x\cb.
\end{equation*}
It is the expectation of the adjoint state, defined by
$$
\textstyle
a(t, \Xb^u, u) = \nabla_{X_t^u}\op \int_t^1 \frac{w(s)}{2}\Vert u(X_s^u,s)\Vert^2 ds - a r(X_1^u) \cp,
$$
where  $\nabla_{X_t^u}$ denotes the derivative with respect to the value $X_t^u$ of the process at time $t$, keeping the trajectory of the Brownian motion $(B_s)_{s\geq t}$ fixed.
It can be computed by solving the backward ODE
\begin{multline}
\label{eq:adjoint-ODE} \textstyle
\frac{d}{dt}a(t, \Xb^u, u) = - \nabla_{X_t^u}\op\frac{w(t)}{2}\Vert u(X_t^u,t)\Vert^2\cp
\\ - \op\nabla_{X_t^u}(b(X_t^u,t) + \gamma(t) u(X_t^u,t))\cp^\top a(t,\Xb^u,u), 
\end{multline}
with initial condition $a(1, \Xb^u, u) = -a\nabla r(X_1^u)$.

Adjoint matching relies on two key ideas. First, using the adjoint state in the equation~\eqref{eq:formula_u_grad_J} gives 
\begin{equation}
\label{eq:formula_optimal_control_adjoint}
u^* (x,t)  = -\frac{\gamma(t)}{w(t)}\E[a(t,\Xb^{u^*}, u^*)|X_t =x].
\end{equation}
Secondly, for the optimal control $u^*$, the term $\nabla_{X_t^u}(\gamma(t) u(X_t^u,t))^\top a(t,\Xb^u,u) + \nabla_{X_t^u}\op \frac{w(t)}{2}\Vert u(X_t^u,t)\Vert^2\cp$ in~\eqref{eq:adjoint-ODE} has zero expectation.
Having a conditional expectation in~\eqref{eq:formula_optimal_control_adjoint}, we can try to enforce this equality by using a square-norm matching trick similar to \citet{vincentConnectionScoreMatching2011}. This leads to the following loss, where $\td a$ is the lean adjoint state, obtained by dropping the zero-expectation terms,
\begin{align}
\L_{\textnormal{WAM}}&(u) := \textstyle  \E_{\Xb \sim p^{\bar u}}\ob \frac{1}{2}\int_0^1\left\Vert u(X_t, t) + \frac{\gamma(t) \td a (t,\Xb)}{\textcolor{red}{w(t)}} \right\Vert^2 dt\cb\notag \\
& \quad\quad (\bar u = \st(u)) \label{eq:adjoint-matching-weighted}\\
\textnormal{where} & \quad \frac{d}{dt} \td a(t, \Xb)  = -\nabla_xb(X_t,t)^\top\td a(t,\Xb),\label{eq:adjoint-matching-weighted:adjoint_ODE}\\
& \quad\, \td a(1,\Xb) = -a\nabla r(X_1).\notag
\end{align}
Our adaptation of adjoint matching turns out to be straightforward, with the extra time weight $\textcolor{red}{w(t)}$ that only appears at one place  in the loss.

\textbf{Proper formulation of the ``stopgrad''.} The stopgrad is convenient both as a notation and for implementing the algorithm in an autodiff framework.
However, when taking the derivative in $u$ of~\eqref{eq:adjoint-matching-weighted}, the mathematical object that we consider is in fact the function of two variables $(u,v)$
$$
\textstyle
\hat \L_{\textnormal{WAM}}(u,v) = \E_{\Xb^v \sim p^{v}}\ob \scriptstyle\frac{1}{2}\int_0^1\left\Vert u(X_t, t) + \frac{\gamma(t) \td a (t,\Xb^v)}{w(t)} \right\Vert^2 dt\cb,
$$
that we differentiate with respect to $u$ then take $v=u$: $\frac{\delta}{\delta u}\L_{\textnormal{WAM}}(u) := \frac{\delta}{\delta u}\hat\L_{\textnormal{WAM}}(u,v)|_{v=u}$.
While commonly used and accepted in the literature, doing gradient descent on an objective with a stopgrad gives no guarantee of decreasing the objective or of converging to a fixed point.
We can, however, characterize the critical points of the objective, where the derivative is zero.

\begin{proposition}
\label{prop:critical_point}
The only critical point $\frac{\delta}{\delta u}\L_{\textnormal{WAM}}(u)=0$ is $u^*$, the optimal control of the weighted SOC problem~\eqref{eq:soc-weighted}.
\end{proposition}
The derivations above are detailed, and this result is proven, in Appendix~\ref{appendix:weighted_adjoint_matching_algo}.
Note also that the critical point is characterized through the derivative with respect to the control $u$ itself, i.e., the functional derivative defined in Appendix~\ref{appendix:weighted_adjoint_matching_algo}.
In practice, $u$ is parametrized by a neural network, $u=u_\theta$ with $\theta \in \R^p$, and so is the loss, $\L_\theta := \L_{\textnormal{WAM}}(u_\theta)$.
The chain rule gives $\nabla_\theta \L_\theta = (\nabla_\theta u_\theta)^* \frac{\delta}{\delta u}\L_{\textnormal{WAM}}(u)$, with $\nabla_\theta u_\theta$ the Jacobian of $\theta\mapsto u_\theta$ and $(\nabla_\theta u_\theta)^* : L_2(\R^d\times[0,1]) \rightarrow \R^p$ its adjoint.
Hence the condition $\nabla_\theta \L_\theta =0$ is weaker than $\frac{\delta}{\delta u}\L_{\textnormal{WAM}}(u_\theta) = 0$: $(\nabla_\theta u_\theta)^*$ is not injective, since a finite-dimensional parameter $\theta \in \R^p$ cannot move the control in every direction.

\textbf{Parametrization of the control.}
The pretrained flow matching model is a deep neural network $v^\textnormal{base} = v_{\theta_0}$ with parameters $\theta_0 \in \R^p$.
We choose to parametrize the control $u$ as 
$$
u_\theta(x,t) = \frac{2}{\gamma(t)}(v^\ft_\theta(x,t)- v^\base(x,t)),
$$
with $ v^\ft_\theta(x,t) = v_\theta(x,t)$ and we initialize with $\theta = \theta_0$, i.e., $u_\theta = 0$.
The loss is therefore
\begin{align}
\label{eq:loss_unstable}
&\L_{\textnormal{WAM}}(\theta) := \E_{\Xb \sim p^{\bar{u_\theta}}}\Big[
\\
&\textstyle\frac{1}{2}\int_0^1\scriptstyle\Big\Vert 2\gamma(t)^{-1}(v^\ft_\theta(X_t,t)- v^\base(X_t,t)) 
+ \frac{\gamma(t) \td a (t,\Xb)}{w(t)} \Big\Vert^2  dt\Big].\notag
\end{align}
For the classical choice of schedules $\alpha_t = t$ and $\beta_t = 1-t$, $\gamma(t) = \sqrt{\frac{2(1-t)}{t}}$, so the factors $\gamma(t)$ and $\gamma(t)^{-1}$ blow up as $t \to 0$ and $t \to 1$.
Consequently, we observed numerical instability when using~\eqref{eq:loss_unstable}.
It turns out that this can be solved by adding a time-dependent weight $\lambda(t)$.

\textbf{Adding an additional time-dependent weight for numerical stability.} In the time integral of the loss~\eqref{eq:adjoint-matching-weighted}, we can insert a positive time-dependent weight $\lambda(t) > 0$ without changing the critical point, leading to the loss: 
\begin{multline}
\label{eq:adjoint-matching-weighted-stab}
\L_{\textnormal{WAM}}^\lambda(u) := \\\E_{\Xb \sim p^{\bar u}}\ob\textstyle\frac{1}{2}\int_0^1\lambda(t)\left\Vert u(X_t, t) + \frac{\gamma(t) \td a (t,\Xb)}{w(t)} \right\Vert^2  dt\cb.
\end{multline}

For the parametrization of the control above, taking $\lambda(t) = 2(1-t)t$ cancels the exploding factors $\gamma(t)$ and $\gamma(t)^{-1}$ of~\eqref{eq:loss_unstable}:
\begin{multline*}
\L_{\textnormal{WAM}}^\lambda(\theta) = \E_{\Xb \sim p^{\bar u_\theta}}\Big[
\\
\textstyle 2\int_0^1\scriptstyle\Big\Vert t(v^\ft_\theta(X_t,t)- v^\base(X_t,t)) 
+ (1-t)\frac{\td a (t,\Xb)}{w(t)} \Big\Vert^2  dt\Big].
\end{multline*}

The algorithm is then obtained by discretizing the generative SDE~\eqref{eq:soc-weighted:SDE} and the adjoint ODE~\eqref{eq:adjoint-matching-weighted:adjoint_ODE} and estimating the expectation over a batch of $M\in \mathbb{N}$ trajectories.
It is summarized in Algorithm~\ref{alg:weighted-adjoint-matching} in Appendix~\ref{appendix:alg:weighted-adjoint-matching}.

\section{Experiments}
\label{sct:experiments}

We ran several experiments to demonstrate how weighted adjoint matching achieves a better tradeoff between reward maximization and balance.%
\footnote{Experiment details and additional experiments can be found in Appendix~\ref{appendix:additional_experiments}.}

The optimal control of the weighted SOC problem~\eqref{eq:soc-weighted} is invariant under the rescaling $(w,a) \leftarrow \op c w, c a\cp$ for $c > 0$, so the value of $a$ is only meaningful relative to~$w$.
Adjoint matching corresponds to choosing $w(t) = 1$ constant, while weighted adjoint matching uses a varying $w$.
We fix the normalization $w(1)=1$ for all our weights.
Since different weights $w$ still regularize differently at the same $a$, we plot the quantities of interest against the achieved average reward $R = \E[r(X_1^{u_\theta})]$ rather than against $a$.

\subsection{Gaussian mixture}
\label{sct:exp:GM}
\begin{figure}[h]
    \centering
    \captionsetup[subfigure]{justification=centering}
    \subfloat[][Exponential weight $w_\textrm{exponential}(t)$.]{
    \includegraphics[width=\linewidth]{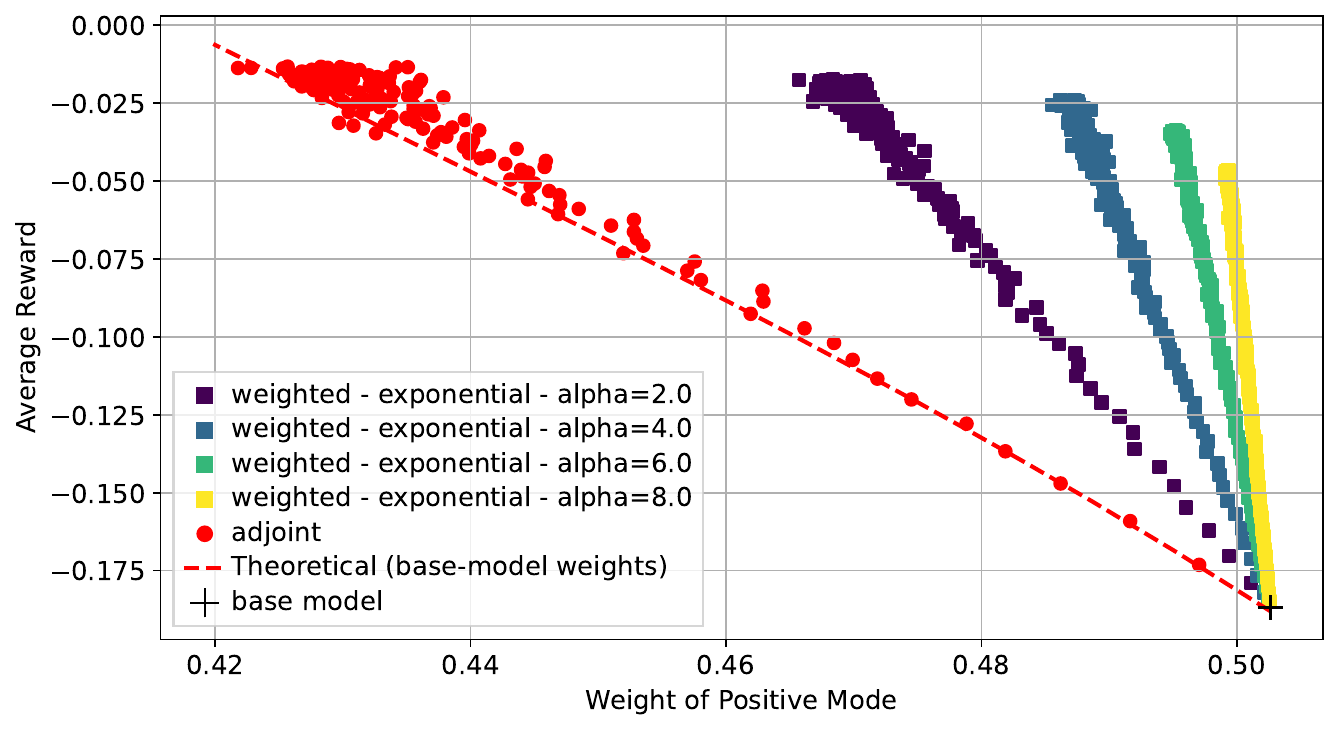}
    }\\
    \subfloat[][Piecewise constant weight $w_\textrm{steps}(t)$.]{
    \includegraphics[width=\linewidth]{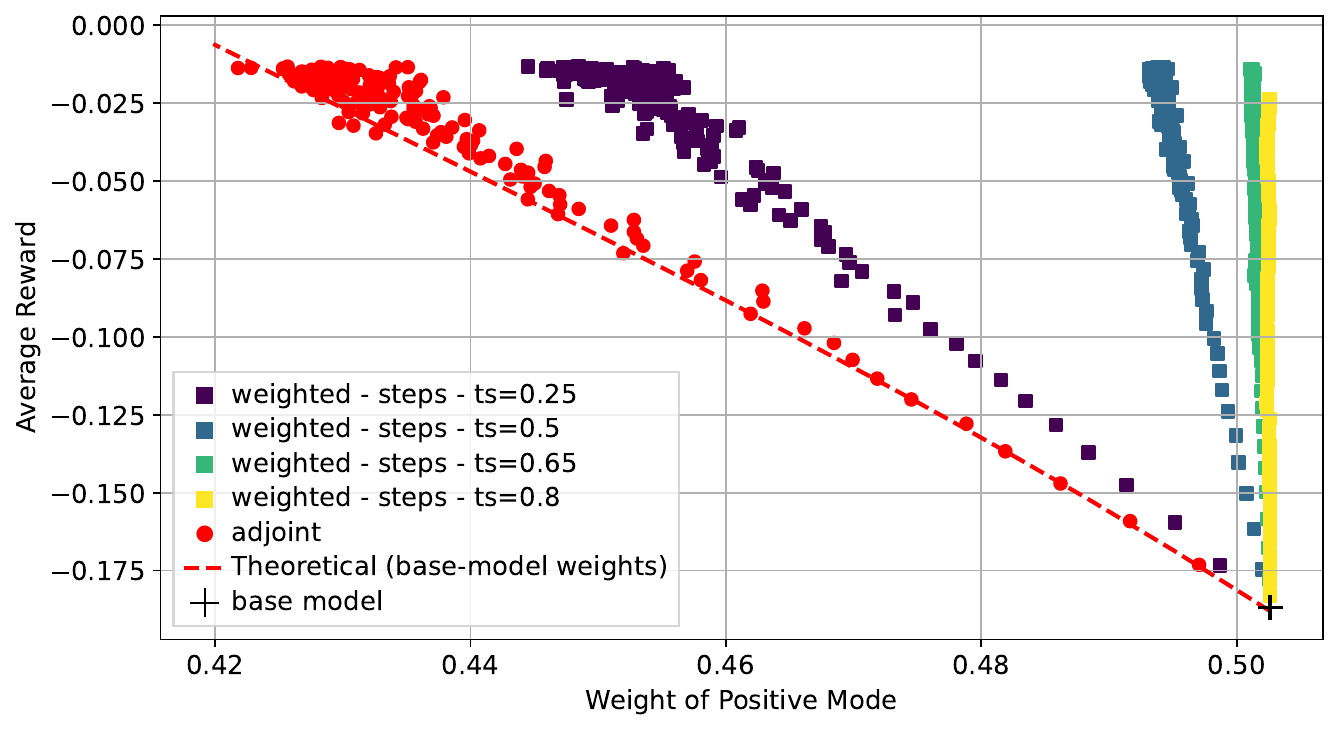}
    }
    \caption{Average reward with respect to weight of positive mode for adjoint matching and weighted adjoint matching, using different functions $t\mapsto w(t)$.}
	\label{fig:reward_weight_GM}
\end{figure}

We use the Gaussian mixture defined in Section~\ref{sct:introduction}
$$
\textstyle
p_\target(x) = \frac{1}{2}\N(x;-\mu,\sigma_1^2I) + \frac{1}{2}\N(x;\mu,\sigma_2^2I).
$$
We train a flow matching model to sample from $p_\target$, then use both adjoint matching \cite{domingo-enrichAdjointMatchingFinetuning2025} and our weighted adjoint matching to finetune the model with the reward $r(x) = -\frac{\Vert x_2\Vert^2}{2}$, which encourages samples to be closer to the first axis.
We take $\mu =(2,0)$, $\sigma_1^2 = 0.25$ and  $\sigma_2^2 = 0.5$ and different values of $a$.
For weighted adjoint matching, we use different types of weighting function $w$: an exponential function $w_\textrm{exponential}(t) = e^{\alpha(1-t)}$, $\alpha >0$, and  a piecewise constant function 
$
w_\textrm{steps}(t) = \left\{\begin{smallmatrix}
    10^5 & \text{if } t<t_s \\
    1 & \text{if } t\geq t_s
\end{smallmatrix}\right.
$, 
$t_s \in (0,1)$. Both regularize more the beginning of the process to try to preserve balance between the two modes; $w_\textrm{steps}(t)$ takes the value $10^5$ on $[0,t_s)$ to force the finetuned model to be almost identical to the base model on this interval.

In Figure~\ref{fig:reward_weight_GM}, we plot empirical estimates of $\kappa^+ = \E[\1_{X_1^{u_\theta} \geq 0}]$ and  $R = \E[r(X_1^{u_\theta})]$ for different runs of the algorithms at different values of $a$, as well as the theoretical curve for the tilted distribution~\eqref{eq:tilted_distribution}  derived in Appendix~\ref{appendix:theory_GM} as a red dotted line.%
\footnote{For the theoretical curve, we use the mode weights $\kappa_0^1,\kappa_0^2$ estimated for the pretrained base model instead of $\frac{1}{2}$.}
For the tilted distribution $p_a$, a higher reward comes with a more unbalanced distribution.
We observe that adjoint matching follows the theoretical curve\footnote{This is consistent with the claim of \cite{domingo-enrichAdjointMatchingFinetuning2025} that it samples from the tilted distribution~\eqref{eq:tilted_distribution}.}, meaning that higher reward is achieved at the price of an unbalanced distribution.
For high values of the reward, corresponding to high values of $a$, the distribution is even more unbalanced than the theory predicts, which could be a consequence of optimization artifacts. 

On the other hand, weighted adjoint matching with both $w_\textrm{exponential}$ and $w_\textrm{steps}$ gives a better tradeoff between reward and balance, with outputs in the upper right corner. With $w_\textrm{steps}$, the reward is increased without changing the balance between the modes for $t_s \in \{0.5,0.65,0.8\}$, while for $t_s = 0.25$, it remains close to adjoint matching.
This is explained by the fact that at time $0.25$, the two modes are not yet separated, and mass can still move between them.

\subsection{MNIST}
\label{sct:exp:MNIST}
\begin{figure}
    \centering
    \includegraphics[width=\linewidth]{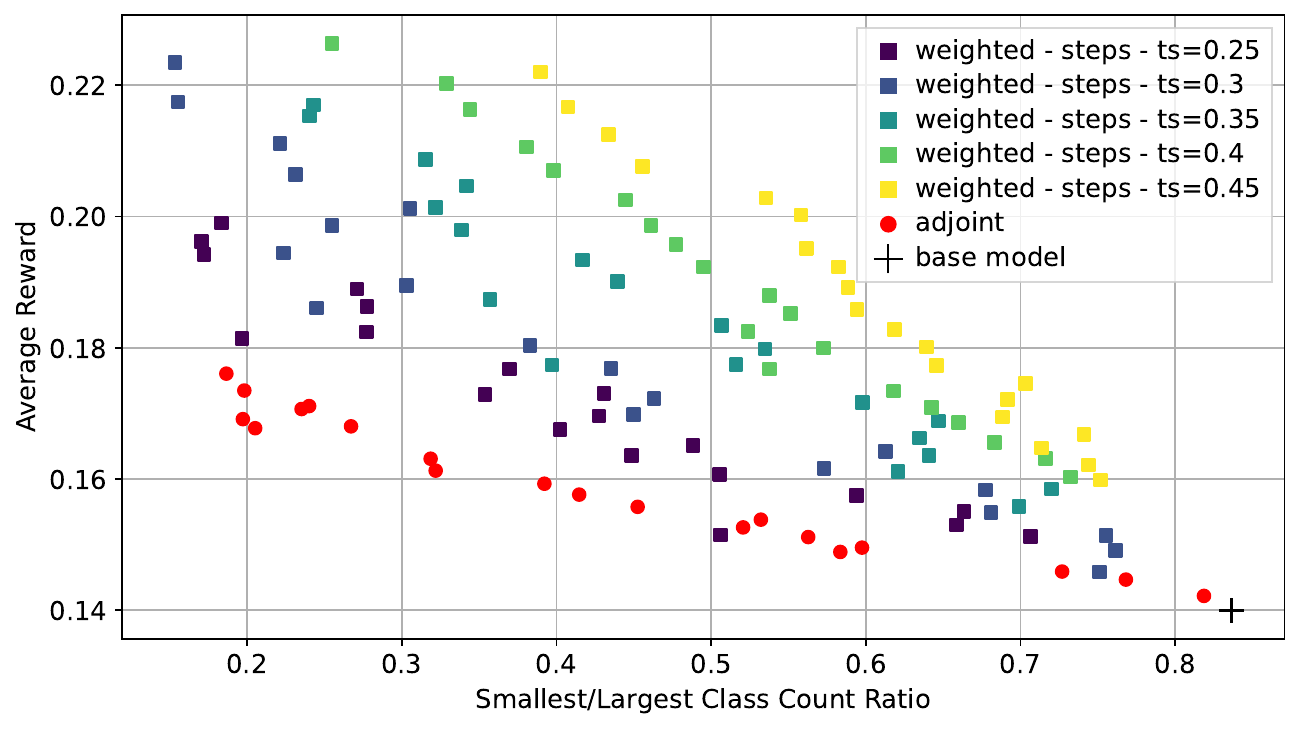}
    \caption{Average brightness reward on the MNIST dataset with respect to class count ratio $\frac{n_{\min}}{n_{\max}}$.}
    \label{fig:reward_by_smallest_largest_class_ratio_steps}
\end{figure}

We train a flow matching model on MNIST \citep{lecunMNISTDatabaseHandwritten1998}. We do not condition on the labels, as we use them as a proxy to evaluate how diversity is preserved.
The reward is the average luminosity: for a digit represented as $x\in [0,1]^{d}$ with $d = 28\times 28 = 784$, $r(x) = \frac{1}{d}\sum_i x_i$.
We generate $50\,000$ samples for each finetuned model and classify them with a pretrained classifier. 
Denoting $n_{\max}$ the number of elements in the largest class and $n_{\min}$ in the smallest class, we evaluate diversity by computing the ratio $\frac{n_{\min}}{n_{\max}}$.
For the pretrained models, this ratio is close to that of the MNIST dataset, which is roughly $0.80$.\footnote{Class size in training set from $5421$ (5's) to $6742$ (1's).}
Results are plotted in Figure~\ref{fig:reward_by_smallest_largest_class_ratio_steps}.
Distributions of all digits, for the different methods at an achieved reward of $0.16$ and $0.18$, are given in Tables~\ref{tab:exp-MNIST_class_percentages_0.16}~\&~\ref{tab:exp-MNIST_class_percentages_0.18} in the Appendix.
Overall, weighted adjoint matching achieves a better tradeoff (closer to the upper right corner) between reward and class balance.
For example, for a reward of $0.176$, adjoint matching generates $18.0\%$ of zeros and only $3.4\%$ of ones, while weighted adjoint matching ($t_s=0.45$) reaches a reward of $0.180$ with $10.5\%$ of zeros and $9.1\%$ of ones.

\section{Discussion}
\label{sct:discussion}
\textbf{Wasserstein regularization.}
Another route would be to drop the KL regularization of~\eqref{eq:reward_objective_KL} altogether and to regularize with the Wasserstein distance $\frac{1}{a} W_2^2(p, p_\target)$ instead.
This penalizes how far each sample moves from its initial position, and therefore limits the transfer of mass between distant modes.
This problem has an explicit solution: first sample $x \sim p_\target$, then apply the proximal operator ${x \mapsto \argmax_y \, r(y) - \frac{1}{a}\Vert x - y\Vert^2}$ (see, e.g., \citep[][]{blanchetQuantifyingDistributionalModel2019, gaoDistributionallyRobustStochastic2023, sinhaCertifyingDistributionalRobustness2020, moitraTractabilityLandscapeDiffusion2026}).
Every sample takes one proximal step on the reward, which roughly preserves the mode weights, but may drive samples out of the data manifold.
For example, on MNIST with the brightness reward, the gradient of $r$ is the constant white image: the background turns gray instead of the digits getting brighter.

\textbf{What is a good reward?}  Some rewards, such as the average luminosity on MNIST of Section~\ref{sct:exp:MNIST}, are very easy to define and to interpret, but have two caveats: they can be increased by going in a direction orthogonal to the data manifold (e.g., making the black background of MNIST digits turn gray) and are not meaningful for comparing samples far apart (dark eights are brighter than bright ones).
We believe that this cannot be easily solved by changing the reward.
Rewards are contextual, and the values have little meaning by themselves: one needs to add the knowledge of the base distribution.
KL regularization addresses the first caveat, since it forces reward maximization to take place on the data manifold, but not the second: it has no notion of distance within the support, so mass can move between distant modes.
Our method, using the link between spatial resolution and time in diffusion models, addresses the second caveat. 

\textbf{Future work.} We have focused on the case of unconditional generative modeling and simple datasets for which balance could be evaluated precisely.
Future work could be dedicated to scaling this work to conditional flow matching models on large image datasets beyond MNIST and to other types of data used in practical applications.

\subsection*{Acknowledgments}
The authors thank Marylou Gabrié, Benjamin Dupuis, Tyler Farghly, and Adam Perbost for insightful discussions related to this work.
The authors are grateful to the CLEPS infrastructure from the Inria of Paris for providing resources and support
This work has received support from the French government, managed by the National Research Agency, under the France 2030 program with the reference ``PR[AI]RIE-PSAI'' (ANR-23-IACL-0008).

\subsection*{AI use statement}

In this work, we used generative AI tools to check equations and proofs of the results, and suggest corrections when appropriate.
AI was also used to write and review the code of our experiments.
The core implementation of Algorithm~\ref{alg:weighted-adjoint-matching} was written by the authors, then reviewed and improved with AI assistance.
AI tools played a larger role in the rest of the code: implementing neural nets, suggesting optimization hyperparameters, writing scripts to launch training runs and to save and retrieve checkpoints and results, generating figures from experimental data, and consolidating the codebase.
Additionally, we used generative AI tools to proofread and improve the writing of the manuscript (including correcting grammar mistakes, suggesting rephrasing to improve clarity and fluency, brainstorming a title and keywords, drafting some paragraphs from bullet points, moving parts of the text to the appendix), to identify relevant literature, and to source or search for information (e.g., to answer background mathematical questions).
Figure~\ref{fig:illustration_algo}, which is purely illustrative, was created using Claude Design from a hand drawing by the authors.
We also used AI to brainstorm and explore some research directions that do not appear in the paper.
We have reviewed all AI-assisted work.
We take responsibility for the final content of this work, including text, claims, or artifacts produced with the aid of generative AI.

\printbibliography[heading=aistats]

\clearpage
\appendix
\thispagestyle{empty}
\onecolumn
\aistatstitle{Weighted Adjoint Matching: Finetuning Flow Matching Models with Time-Dependent Control Cost \\ Supplementary Materials}

\section{Gaussian mixture with KL-regularized objective}
\label{appendix:theory_GM}
We compute the tilted distribution $p_a$ of a two-component Gaussian mixture in $\R^2$ and show that it does not preserve the weights of the modes.
We define $p_\target$ as a mixture of two modes
$$
p_\target(x) = \kappa_0^1\N(x;\mu_1,\sigma_1^2I) + \kappa_0^2\N(x;\mu_2,\sigma_2^2I) = \kappa_0^1 (2 \pi \sigma_1^2)^{-1}e^{-\frac{\Vert x - \mu_1 \Vert^2}{2 \sigma_1^2}} + \kappa_0^2 (2 \pi \sigma_2^2)^{-1}e^{-\frac{\Vert x - \mu_2 \Vert^2}{2 \sigma_2^2}},
$$
with $\mu_1 = (-m,0)$, $\mu_2 = (m,s)$, $m,s\in\R$, $\sigma_1,\sigma_2 >0$ and $\kappa_0^1,\kappa_0^2 >0$ such that $\kappa_0^1 + \kappa_0^2 = 1$.
For $x = (x_1,x_2)$,  we define the reward $r(x) = -\frac{\Vert x_2\Vert^2}{2}$ to encourage samples to be closer to the first axis. 
The tilted distribution is 
$$
p_a(x) = Z_a^{-1} p_\target(x)e^{ar(x)},
$$
with $Z_a = \E_{p_\target}[e^{ar(x)}]$.
It can be rewritten as:
\begin{align*}
p_a(x)
&= \frac{\kappa_0^1}{Z_a} \frac{\det(\Sigma_{1,a})^{1/2}}{\sigma_1^2} \frac{\exp\op-\frac{\Vert x - \mu_1 \Vert^2_{\Sigma_{1,a}^{-1}}}{2}\cp}{2\pi \det(\Sigma_{1,a})^{1/2}} +
\frac{\kappa_0^2}{Z_a} \frac{\det(\Sigma_{2,a})^{1/2}\exp\op -\frac{as^2}{2(1+a\sigma_2^2)}\cp}{\sigma_2^2} \frac{\exp\op-\frac{\Vert x - \mu_{2,a} \Vert^2_{\Sigma_{2,a}^{-1}}}{2}\cp}{2\pi \det(\Sigma_{2,a})^{1/2}}\\
&=
\kappa_a^1 \N(x;\mu_1,\Sigma_{1,a}) + \kappa_a^2\N(x;\mu_{2,a},\Sigma_{2,a})
\end{align*}
with $\Vert x\Vert_A^2 = x^\top Ax$ and for $i\in \{1,2\}$
$$
\Sigma_{i,a} =
\begin{bmatrix}
 \sigma_i^2 & 0\\
 0 & \frac{\sigma_i^2}{1 + a \sigma_i^2}
\end{bmatrix}, 
\quad 
\mu_{2,a} = \op m,\frac{s}{1+a\sigma_2^2}\cp.
$$
We recognize a mixture of two Gaussian distributions with $\kappa^1_a$ the weight of the first mode and $\kappa^2_a$ the weight of the second mode equal to
\begin{align*}
\kappa^1_a &= \frac{\kappa_0^1}{Z_a} \frac{\det(\Sigma_{1,a})^{1/2}}{\sigma_1^2} ,\\
\kappa^2_a &=\frac{\kappa_0^2}{Z_a} \frac{\det(\Sigma_{2,a})^{1/2}}{\sigma_2^2} \exp\op-\frac{as^2}{2(1+a\sigma_2^2)}\cp,
\end{align*}
from which we deduce that
\begin{align*}
\kappa_a &= \frac{\kappa^1_a}{\kappa^2_a} = \kappa_0\sqrt{\frac{1 + a \sigma_2^2}{1 + a \sigma_1^2}}\exp\op\frac{as^2}{2(1+a\sigma_2^2)}\cp, \quad \kappa_0 = \frac{\kappa^1_0}{\kappa^2_0},\\
\kappa^1_a &= \frac{\kappa_a}{1+\kappa_a} = \frac{\kappa^1_0\sqrt{1+a\sigma_2^2}\exp\op\frac{as^2}{2(1+a\sigma_2^2)}\cp}{\kappa^2_0\sqrt{1+a\sigma_1^2}+\kappa^1_0\sqrt{1+a\sigma_2^2}\exp\op\frac{as^2}{2(1+a\sigma_2^2)}\cp},\\
\kappa^2_a &= \frac{1}{1+\kappa_a} = \frac{\kappa^2_0\sqrt{1+a\sigma_1^2}}{\kappa^2_0\sqrt{1+a\sigma_1^2}+\kappa^1_0\sqrt{1+a\sigma_2^2}\exp\op\frac{as^2}{2(1+a\sigma_2^2)}\cp}.
\end{align*}

In particular, the ratio depends on $a$, and as $a \rightarrow + \infty$, the tilted distribution converges to a mixture of two degenerate Gaussian distributions (a projection on the first axis defined by the equation $x_2 = 0$) with a weight ratio
$$
\kappa_\infty = \lim_{a\rightarrow + \infty} \kappa_a = \kappa_0\frac{\sigma_2}{\sigma_1} \exp\op\frac{s^2}{2\sigma_2^2}\cp .
$$
We also define $\kappa_a^- = \E_{p_a}[\1_{x_1 \leq 0}]$ and $\kappa_a^+ = \E_{p_a}[\1_{x_1 > 0}] = 1 - \kappa_a^-$, which will be close to $\kappa_a^1$ and $\kappa_a^2$ if the modes are well separated, but easier to estimate empirically:
\begin{align*}
\kappa_a^- &= \E_{p_a}[\1_{x_1 \leq 0}] \\
&= \kappa_a^1\E_{x \sim \N(\mu_1, \Sigma_{1,a})}[\1_{x_1 \leq 0}] + \kappa_a^2\E_{x \sim \N(\mu_{2,a}, \Sigma_{2,a})}[\1_{x_1 \leq 0}] \\
&= \kappa_a^1\Phi\op\frac{m}{\sigma_1}\cp+ \kappa_a^2 \Phi\op\frac{-m}{\sigma_2}\cp,\\
\kappa_a^+ &= \kappa_a^1\Phi\op\frac{-m}{\sigma_1}\cp+ \kappa_a^2 \Phi\op\frac{m}{\sigma_2}\cp,
\end{align*}
where $\Phi(x) = \frac{1}{\sqrt{2\pi}}\int_{-\infty}^xe^{-t^2/2}dt$ is the cumulative distribution function (CDF) of the standard normal distribution.
Finally, we compute the optimal reward $R_a$ achieved by $p_a$:
\begin{align*}
R_a &= \E_{p_a}[r(X)] \\
&= \kappa_a^1 \E_{\N(\mu_1, \Sigma_{1,a})}[r(X)] + \kappa_a^2 \E_{\N(\mu_{2,a}, \Sigma_{2,a})}[r(X)] \\
&= -\frac{1}{2}\op \kappa_a^1 \frac{\sigma_1^2}{1 + a \sigma_1^2} + \kappa_a^2 \op\frac{\sigma_2^2}{1 + a \sigma_2^2} + \op\frac{s}{1+a\sigma_2^2}\cp^2\cp\cp.
\end{align*}

\section{Diffusion and flow matching models: forward noising and reverse processes}
\label{appendix:intro_diffusion}
In this section, we give more details on the derivation of Section~\ref{sct:intro_diffusion}.

\paragraph{Forward process of diffusion models.} The Gaussian noise corruption process \citep[see, e.g.,][]{songScorebasedGenerativeModeling2021,karrasElucidatingDesignSpace2022} is defined by
\begin{equation*}
\tag{\ref{eq:sde_forward_gen}}
\left\{
\begin{array}{l}
d\fw X_\tau =  -f(\tau)\fw X_\tau d\tau + g(\tau)dB_\tau, \\
\fw X_0 = X,
\end{array}\right.
\end{equation*}
where $f,g : \R \mapsto\R_+$ are continuous functions.
To derive the closed-form solution of~\eqref{eq:sde_forward_gen}, denote $Y_\tau = a(\tau)^{-1}\fw X_\tau$, with $a(\tau) = \exp\op-\int_0^\tau f(u)du\cp$.
Itô's formula \citep[see, e.g.,][]{legallBrownianMotionMartingales2016} gives
$$
d Y_\tau = f(\tau)a(\tau)^{-1}\fw X_\tau d\tau +  a(\tau)^{-1}d\fw X_\tau = \frac{g(\tau)}{a(\tau)}dB_\tau,
$$
hence
$$
Y_\tau = a(0)^{-1}\fw X_0 + \int_0^\tau\frac{g(u)}{a(u)}dB_u = X + \int_0^\tau\frac{g(u)}{a(u)}dB_u,
$$
leading to the closed-form solution,
$$
\fw X_\tau=  a(\tau) X +  a(\tau)\int_0^\tau\frac{g(u)}{a(u)}dB_u.
$$
In particular, for each time $\tau$, the joint distribution $\L((\fw X_0,\fw X_\tau))$ satisfies 
$$
\L\op(\fw X_0,\fw X_\tau)\cp = \L\op(X, a(\tau) X + b(\tau) Z)\cp
$$
with $Z\sim \N(0,I)$ independent from $X$ and $b(\tau) = a(\tau)\sqrt{\int _0^\tau \frac{g(u)^2}{a(u)^2}du}$.

\paragraph{Backward process.} The idea behind diffusion models is to fix a large $T>0$, such that $\fw X_T$ is close to a standard normal distribution, and then to find a process that has the same marginal laws as $(\fw X_{T-t})_{t\in[0,T]}$ to map a Gaussian to the target distribution. 
The marginal laws of $\fw X_\tau$ are characterized by the Fokker--Planck equation \citep[see, e.g.,][]{riskenFokkerPlanckEquationMethods1996,bogachevFokkerPlanckKolmogorov2015} that governs the evolution of the density $\fw p_\tau$ of $\fw X_\tau$
$$
\partial_\tau \fw p_\tau = - \nabla\cdot(-f(\tau)x \fw p_\tau) + \frac{1}{2}g(\tau)^2\Delta \fw p_\tau.
$$
Reversing time with $t = T -\tau$, $\rv p_t = \fw p_{T-t}$, $\rv f(t) = f(T-t), \rv g(t) = g(T-t)$, gives
$$
\partial_t \rv p_t = - \nabla\cdot(\rv f(t)x\rv p_t) - \frac{1}{2}\rv g(t)^2\Delta \rv p_t,
$$
but here the term in front of the Laplacian $\Delta$ is negative, so this cannot be read as the Fokker--Planck equation of an SDE.
The trick is to introduce $\gamma: \R \rightarrow \R^+$, and use $\Delta\rv p_t = \nabla \cdot (\rv p_t\nabla\log \rv p_t)$ to get 
$$
\partial_t \rv p_t=  - \nabla\cdot\op\op\rv f(t)x + \frac{\gamma(t)^2 + \rv g(t)^2}{2}\nabla\log \rv p_t\cp\rv p_t\cp + \frac{\gamma(t)^2}{2}\Delta \rv p_t
$$
With a positive term in front of $\Delta$, this is the Fokker--Planck equation of the process $\rv X_t$ defined by
\begin{equation*}
\tag{\ref{eq:general_backward_process}}
\left\{
\begin{array}{ll}
d\rv{X}_t &=  \op \rv f(t)\rv X_t +   \frac{\gamma(t)^2 + \rv g(t)^2}{2}\nabla \log \rv p_t(\rv X_t)\cp dt + \gamma(t)dW_t,  \\
\rv{X}_0 &= \fw X_T, 
\end{array}\right.
\end{equation*}
with $W_t$ a Brownian motion.
Hence this process satisfies $\forall t \in [0,T], \L(\rv X_t) = \L(\fw X_{T-t})$.

Note that the Fokker--Planck equation can only ensure the equality of the marginal laws. 
However, a result by \citet{haussmannTimeReversalDiffusions1986} proves that taking $\gamma(t)^2= \rv g(t)^2$ also preserves the path distribution $\L((\rv X_t)_{t\in[0,T]}) = \L((\fw X_{T-t})_{t\in[0,T]})$.

\paragraph{Flow matching and the memoryless noise schedule.} \citet{domingo-enrichAdjointMatchingFinetuning2025} generalized flow matching~\eqref{eq:flow-matching-ODE} by adding a diffusion coefficient $\gamma: [0,1] \mapsto \R$ as 
\begin{equation*}
\tag{\ref{eq:flow-matching-SDE}}
\left\{
\begin{array}{ll}
     dX_t & = b^\gamma(X_t,t)  dt + \gamma(t)dB_t\\
     X_0 & \sim \N(0,I),
\end{array}
\right. \quad \textnormal{ with } b^\gamma(x,t) = v(x,t) + \frac{\gamma(t)^2}{2}\nabla\log p_t(x) 
\end{equation*}
They give the closed form of the velocity field (in their Appendix~B.4)
\begin{equation*}
\tag{\ref{eq:relation_velocity_score}}
v(x,t) = \frac{\dot \alpha_t}{\alpha_t}x + \beta_t\op\frac{\dot\alpha_t}{\alpha_t}\beta_t - \dot\beta_t\cp \nabla\log p_t(x),
\end{equation*}
hence 
$$
b^\gamma(x,t) =   \frac{\dot \alpha_t}{\alpha_t}x + \op \beta_t\op\frac{\dot\alpha_t}{\alpha_t}\beta_t - \dot\beta_t\cp + \frac{\gamma(t)^2}{2}\cp\nabla\log p_t(x).
$$
Using this expression, we see that~\eqref{eq:flow-matching-SDE} is just the backward SDE~\eqref{eq:general_backward_process} for a specific choice of $f(\tau)$ and $g(\tau)$.
Indeed, taking 
\begin{align*}
f(\tau) &= - \frac{d}{d\tau}\log \alpha_{1-\tau} = \frac{\dot \alpha_{1-\tau}}{\alpha_{1-\tau}} \\
g(\tau) &= \alpha_{1-\tau} \sqrt{\frac{d}{d\tau}\op\alpha_{1-\tau}^{-2}\beta_{1-\tau}^2\cp} 
= \sqrt{2 \beta_{1-\tau} \op\frac{\dot\alpha_{1-\tau}}{\alpha_{1-\tau}}\beta_{1-\tau}-\dot\beta_{1-\tau} \cp},
\end{align*}
gives
$$
a(\tau) = \exp\op-\int_0^\tau f(u)du\cp = \exp\op-\int_0^\tau - \frac{d}{du}\log \alpha_{1-u} du\cp = \frac{\alpha_{1-\tau}}{\alpha_1} = \alpha_{1-\tau}
$$
and
$$
b(\tau) 
= a(\tau)\sqrt{\int _0^\tau \frac{g(u)^2}{a(u)^2}du}
= \alpha_{1-\tau} \sqrt{\int _0^\tau \frac{d}{du}\op\alpha_{1-u}^{-2}\beta_{1-u}^2\cp du} = \alpha_{1-\tau} \sqrt{\alpha_{1-\tau}^{-2}\beta_{1-\tau}^2 - \alpha_{1}^{-2}\beta_{1}^2} = \beta_{1-\tau}.
$$
The forward process $\fw X_\tau$ has therefore marginals $\L(a(\tau)X + b(\tau)Z)$, $Z\sim \N(0,I)$ independent from $X$, going from data $\tau =0, a(0) = \alpha_1 =1, b(0) = \beta_1 = 0$ to pure noise at time $T=1$, with $a(1) = \alpha_0 =0, b(1) = \beta_0 = 1$.
The backward process $\rv X_t$~\eqref{eq:general_backward_process}, plugging the values $\rv a(t) = a(1-t) = \alpha_t$, $\rv b(t) = b(1-t) = \beta_t$, $\rv f(t) = \frac{\dot \alpha_t}{\alpha_t}$ and $\rv g(t) = \sqrt{2 \beta_t \op\frac{\dot\alpha_t}{\alpha_t}\beta_t-\dot\beta_t\cp}$, is then simply~\eqref{eq:flow-matching-SDE}.

The flow matching SDE of \citet{domingo-enrichAdjointMatchingFinetuning2025} is therefore just a diffusion model with a particular choice of time schedule.
Moreover, the choice of memoryless noise schedule is 
$$
\gamma(t) = \sqrt{2 \beta_t \op\frac{\dot\alpha_t}{\alpha_t}\beta_t-\dot\beta_t\cp} = \rv g(t),
$$
which is exactly the choice of  $\gamma(t)$ that also conserves the path distribution $\L((\rv X_t)_{t\in[0,T]}) = \L((\fw X_{T-t})_{t\in[0,T]})$ according to \citet{haussmannTimeReversalDiffusions1986}.

\section{Path-KL regularized objective}
\label{appendix:path-KL}

This appendix and Appendix~\ref{appendix:weighted_adjoint_matching_algo} are written for the weighted SOC problem on a time horizon $T>0$, with an arbitrary running cost $f:\R^d\times[0,T]\rightarrow \R$ and terminal cost $g:\R^d\rightarrow \R$,
\begin{align}
\label{eq:soc-weighted-gen}
\min_{u \in \mathcal{U}} \quad & \E \left[ \int_0^T \left( \frac{w(t)}{2}\Vert u(X_t^u, t)\Vert^2 + f(X_t^u, t) \right) dt + g(X_T^u) \right] \\
\text{s.t.} \quad & dX_t^u = \left( b(X_t^u, t) + \gamma(t)\,u(X_t^u, t) \right) dt + \gamma(t)\, dB_t, 
\label{eq:soc-weighted-gen:SDE}
\\
& X_0^u \sim p_0\notag
\end{align}
of which the finetuning problem~\eqref{eq:soc-weighted} of the main text is the particular case $T=1$, $f = 0$ and $g = -ar$.
This section is dedicated to the proof of Proposition~\ref{prop:path-KL}, in the general form of Proposition~\ref{prop:path-KL-gen}, and to a dual reformulation of~\eqref{eq:soc-weighted-gen} in which the regularization applies to the sub-paths $[t,T]$ instead of the sub-paths $[0,t]$.
Both reformulations rely on the following consequence of Girsanov's theorem, which expresses the quadratic control cost on a time interval as the KL divergence between the controlled and the base processes on that interval.
\begin{proposition}
\label{prop:girsanov}
For every $0 \leq s \leq t \leq T$ and every $u \in \mathcal{U}$,
$$
\KL(p^u(X_{[s,t]}|X_s)\Vert p^\base(X_{[s,t]}|X_s)) = \E_{p^u}\ob\frac{1}{2}\int_s^t \Vert u(X_r^u,r)\Vert^2dr \Big| X_s\cb.
$$
\end{proposition}
\begin{proof}
For $s=0$ and $t=T=1$, this is Theorem~2 and Appendix~C.4 of \citet{domingo-enrichAdjointMatchingFinetuning2025}.
The general case follows from the same argument applied to the process $(X_r^u)_{r \in [s,t]}$ started at $X_s$.
\end{proof}
In particular, $K_t$ defined in~\eqref{eq:path-KL} satisfies
\begin{equation}
\label{eq:path-KL-girsanov}
K_t(X_0) = \E_{p^u} \ob\frac{1}{2}\int_0^t \Vert u(X_s^u,s)\Vert^2ds \Big | X_0 \cb.
\end{equation}

\subsection{KL $(0,t)$-path regularization}
\label{appendix:path-KL:past}
\begin{proposition}
\label{prop:path-KL-gen}
Assume that $w$ satisfies~\eqref{eq:weight-integral-form}.
Then the weighted SOC problem~\eqref{eq:soc-weighted-gen} is equivalent to
\begin{equation}
\label{eq:path-KL-regularized-objective-gen}
\max_{u \in \mathcal{U}} \E_{X_0 \sim p_0} \ob\E_{\Xb^u\sim p^u(\cdot|X_0)} \ob
\int_0^T -f(X_t^u, t) dt - g(X_T^u)
\cb - w(T)\KL(p^u(\Xb|X_0) \Vert p^\base(\Xb|X_0)) + \int_0^T K_t(X_0) dw(t) \cb.
\end{equation}
\end{proposition}
Proposition~\ref{prop:path-KL} is the particular case $T=1$, $f=0$, $g=-ar$.
\begin{proof}
Using that the weight function satisfies~\eqref{eq:weight-integral-form},~\eqref{eq:path-KL-girsanov}, and exchanging integrals,
\begin{align*}
\E_{p^u} \ob\int_0^T \frac{w(t)}{2}\Vert u(X_t^u,t)\Vert^2dt \Big | X_0 \cb 
&= w(T) K_T(X_0) - \E \ob \int_{t=0}^T \int_{s=t}^T \frac{1}{2}\Vert u(X_t^u,t)\Vert^2 dw(s)dt\Big| X_0\cb \\
&= w(T) K_T(X_0) - \int_{s=0}^T\E \ob \int_{t=0}^s  \frac{1}{2}\Vert u(X_t^u,t)\Vert^2dt\Big| X_0\cb  dw(s)\\
&= w(T) K_T(X_0) - \int_0^T K_s(X_0)dw(s).
\end{align*}
Plugging this equality into~\eqref{eq:soc-weighted-gen} gives~\eqref{eq:path-KL-regularized-objective-gen}.
\end{proof}

\subsection{KL $(t,T)$-path regularization}
\label{appendix:path-KL:future}
Considering the process between time $t$ and $T$, Proposition~\ref{prop:girsanov} gives
\begin{equation}
\label{eq:path-KL-future}
\td K _t(X_t) := \KL(p^u(X_{[t,T]}|X_t) \Vert p^\textnormal{base}(X_{[t,T]}|X_t)) = \E_{p^u}\ob\frac{1}{2}\int_t^T\Vert u(X_s^u,s)\Vert^2ds\Big| X_t\cb.
\end{equation}
\begin{proposition}
\label{prop:path-KL-future}
Under the assumptions of Proposition~\ref{prop:path-KL-gen}, the weighted SOC problem~\eqref{eq:soc-weighted-gen} is also equivalent to
\begin{equation}
\label{eq:path-KL-regularized-objective-future}
\max_{u \in \mathcal{U}} \mathbb{E}_{X_0 \sim p_0} \ob\mathbb{E}_{\Xb^u\sim p^u(\cdot|X_0)} \ob
-\int_0^T f(X_t^u, t)\, dt - g(X_T^u) - \int_0^T\td K _t(X_t)dw(t)
\cb - w(0)\KL(p^u(\mathbf{X}|X_0) \Vert p^{\text{base}}(\mathbf{X}|X_0))  \cb.
\end{equation}
\end{proposition}
\begin{proof}
With~\eqref{eq:weight-integral-form} and~\eqref{eq:path-KL-future}, as $X_t^u$ is Markovian, we get
\begin{align*}
\E_{p^u} \ob\int_0^T \frac{w(t)}{2}\Vert u(X_t^u,t)\Vert^2dt \Big | X_0 \cb
&= w(0) \td K_0(X_0) + \E_{p^u} \ob \int_{t=0}^T \int_{s=0}^t \frac{1}{2}\Vert u(X_t^u,t)\Vert^2 dw(s)dt\Big| X_0\cb\\
&= w(0) \td K_0(X_0) + \int_{s=0}^T\E \ob \int_{t=s}^T  \frac{1}{2}\Vert u(X_t^u,t)\Vert^2dt \Big| X_0\cb dw(s) \\
&= w(0) \td K_0(X_0) + \int_{s=0}^T\E \ob \E \ob \int_{t=s}^T  \frac{1}{2}\Vert u(X_t^u,t)\Vert^2dt \Big| X_s^u\cb\Big| X_0\cb dw(s) \\
&= w(0) \td K_0(X_0) + \E \ob \int_0^T \td K_s(X_s^u)dw(s)\Big| X_0\cb.
\end{align*}
Plugging this equality into~\eqref{eq:soc-weighted-gen} gives~\eqref{eq:path-KL-regularized-objective-future}.
\end{proof}
Compared to the non-weighted SOC problem, the KL divergence between the paths on $[t,T]$ enters the objective with weight $-dw(t)$, which for a non-decreasing $w$ is negative and therefore adds regularization. 

\paragraph{Example.}
Taking a piecewise constant function $w$
$$
w(t) =
\left\{
\begin{array}{ll}
     1 & \textnormal{if} \quad t< \frac{T}{2}   \\
     2 & \textnormal{if} \quad t > \frac{T}{2}  
\end{array}
\right.
$$
we get the equivalent objective
\begin{multline*} 
\max_{u \in \mathcal{U}} 
\mathbb{E}_{X_0 \sim p_0} \Bigg[\mathbb{E}_{\Xb^u\sim p^u(\cdot|X_0)} \ob
\int_0^T -f(X_t^u, t)\, dt - g(X_T^u) - \KL(p^u(X_{[T/2,T]}|X_{T/2}^u)\Vert p^\base(X_{[T/2,T]}|X_{T/2}^u)) 
\cb \\
-\KL(p^u(\mathbf{X}|X_0) \Vert p^{\text{base}}(\mathbf{X}|X_0))  \Bigg].
\end{multline*}
Divergence from the base distribution on $(T/2,T)$ is penalized twice.

\subsection{Link between the two reformulations}
\label{appendix:path-KL:link}
\begin{proposition}
\label{prop:path-KL-link}
For every $t \in [0,T]$,
$$
K_T(X_0) = \td K_0(X_0), \qquad K_T(X_0) - K_t(X_0) = \E\ob \td K_t(X_t^u)\Big| X_0\cb,
$$
and the regularization terms of~\eqref{eq:path-KL-regularized-objective-gen} and of~\eqref{eq:path-KL-regularized-objective-future} coincide,
$$
-w(T) K_T(X_0) + \int_0^TK_t(X_0)dw(t) = -w(0) K_T(X_0)  - \E\ob \int_0^T\td K_t(X_t)dw(t) \Big | X_0\cb.
$$
\end{proposition}
\begin{proof}
The first equality is immediate from the definitions of $K_T$ in~\eqref{eq:path-KL} and of $\td K_0$ in~\eqref{eq:path-KL-future}.
For the second one, as $X_t^u$ is Markovian
\begin{align*}
K_T(X_0) - K_t(X_0) 
&= \E_{\Xb^u\sim p^u(\cdot|X_0)}\ob\log\op\frac{p^u(X_{[0,T]}|X_0)}{p^\textnormal{base}(X_{[0,T]}|X_0)}\cp\cb
- \E_{\Xb^u\sim p^u(\cdot|X_0)}\ob\log\op\frac{p^u(X_{[0,t]}|X_0)}{p^\textnormal{base}(X_{[0,t]}|X_0)}\cp\cb\\
&= \E_{\Xb^u\sim p^u(\cdot|X_0)}\ob\log\op\frac{p^u(X_{[0,T]}|X_0)}{p^u(X_{[0,t]}|X_0)}\frac{p^\textnormal{base}(X_{[0,t]}|X_0)}{p^\textnormal{base}(X_{[0,T]}|X_0)}\cp\cb \\
&= \E_{\Xb^u\sim p^u(\cdot|X_0)}\ob\log\op\frac{p^u(X_{[t,T]}|X_t)}{p^\textnormal{base}(X_{[t,T]}|X_t)}\cp\cb
= \E\ob\log\op\frac{p^u(X_{[t,T]}|X_t)}{p^\textnormal{base}(X_{[t,T]}|X_t)}\cp\Big| X_0\cb\\
&= \E\ob\E\ob\log\op\frac{p^u(X_{[t,T]}|X_t)}{p^\textnormal{base}(X_{[t,T]}|X_t)}\cp\Big| X_t\cb\Big| X_0\cb = \E[\td K_t(X_t^u)|X_0],
\end{align*}
hence
\begin{align*}
-w(T) K_T(X_0) + \int_0^TK_t(X_0)dw(t) 
&= -w(0) K_T(X_0)  - (w(T) - w(0)) K_T(X_0) + \int_0^TK_t(X_0)dw(t) \\
&= -w(0) K_T(X_0)  + \int_0^T(K_t(X_0)-K_T(X_0))dw(t) \\
&= -w(0) K_T(X_0)  - \E\ob \int_0^T\td K_t(X_t)dw(t) \Big | X_0\cb.
\end{align*}
\end{proof}

\section{The weighted adjoint matching algorithm}
\label{appendix:weighted_adjoint_matching_algo}
As in Appendix~\ref{appendix:path-KL}, this section is written for the weighted SOC problem with an arbitrary running cost $f$ and terminal cost $g$,
\begin{align*}
\tag{\ref{eq:soc-weighted-gen}}
\min_{u \in \mathcal{U}} \quad & \E \left[ \int_0^T \left( \frac{w(t)}{2}\Vert u(X_t^u, t)\Vert^2 + f(X_t^u, t) \right) dt + g(X_T^u) \right] \\
\text{s.t.} \quad & dX_t^u = \left( b(X_t^u, t) + \gamma(t)\,u(X_t^u, t) \right) dt + \gamma(t)\, dB_t, 
\tag{\ref{eq:soc-weighted-gen:SDE}}
\\
& X_0^u \sim p_0\notag
\end{align*}
The finetuning problem~\eqref{eq:soc-weighted} of the main text is the particular case $T=1$, $f=0$ and $g=-ar$.
Note also that this SOC problem and the algorithm that we propose can be defined outside the framework of flow matching.
In~\eqref{eq:soc-weighted-gen}, $b(X_t,t)$ can be any drift and $\gamma(t)\in \R^{d\times d}$ invertible, provided the process is well defined for $u=0$.
When coming back to our original problem, we will use $\gamma(t)$ real-valued and $b = b^\gamma$.

This section is dedicated to the derivation of the weighted adjoint matching algorithm and to the proof of Proposition~\ref{prop:critical_point}.
We roughly follow and adapt the original derivations by \citet{domingo-enrichAdjointMatchingFinetuning2025}, though we differ when necessary or when we believe it makes the exposition clearer.
We try to be as self-contained as possible.
See also \citet{domingo-enrichStochasticOptimalControl2024} for an earlier work that lays the groundwork for the adjoint matching method.
A more recent work \citep{domingo-enrichAdjointMatchingLens2026} proposes a more general Hamiltonian framework for adjoint matching that can partly retrieve our derivation of the weighted adjoint matching.
It does not, however, include the extra time weight $\lambda(t)$ that balances the loss between the different timesteps and gives a more stable algorithm. 

\subsection{Characterization of the optimal control}
\paragraph{Infinitesimal generator.} The infinitesimal generator $L_t$ associated with $$dX_t = b(X_t,t)dt + \gamma(t)dB_t $$ is defined by
$$
L_t f = \lim_{\Delta t \searrow 0 }\frac{\E\ob f(X_{t+\Delta t}) |X_t =x \cb-f(x)}{\Delta t}
$$
for any function $f$ such that the limit exists.
Assuming that $f \in \C^2$, a second-order Taylor expansion gives
\begin{align*}
f(X_{t+\Delta t}) 
&\approx f(X_t) + \langle \nabla f(X_t), b(X_t,t)\rangle\Delta t + \frac{1}{2}(\gamma(t)Z)^\top\nabla^2 f(X_t) \gamma(t)Z \Delta t \\
&\approx f(X_t) + \langle \nabla f(X_t), b(X_t,t)\rangle\Delta t + \frac{1}{2}\tr\op \gamma(t)^\top\nabla^2f(X_t) \gamma(t)ZZ^\top \cp \Delta t ,
\end{align*}
with $Z\sim \N(0,I)$, leading to
$$
L_t f = \langle\nabla f(x), b(x,t)\rangle + \frac{1}{2}\tr\op\gamma(t)^\top\nabla^2f(x) \gamma(t)\cp= \langle\nabla f(x), b(x,t)\rangle + \frac{1}{2}\tr\op\gamma(t)\gamma(t)^\top\nabla^2f(x) \cp,
$$
which we can rewrite as
$$
L_t = \sum_i b_i(x,t)\partial_{x_i} + \frac{1}{2}\sum_{i,j} (\gamma\gamma^\top)_{ij}\partial_{x_i}\partial_{x_j}.
$$
Similarly, for some control $u$, the infinitesimal generator $L_t^u$ associated with $$dX_t^u = (b(X_t^u,t) +\gamma(t) u(X_t^u,t))dt + \gamma(t)dB_t $$ is
$$
L_t^u  = L_t + \sum_i (\gamma(t)u)_i(x,t)\partial_{x_i} 
$$

\paragraph{Cost and value function.} The SOC problem that we consider is~\eqref{eq:soc-weighted-gen}.
The \emph{cost function} is defined, for some control $u$ and $t\in [0,T]$, as
\begin{equation*}
J(u,x,t) = \E_{\Xb \sim p^u}\ob\int_t^T\op\frac{w(s)}{2}\Vert u(X_s,s)\Vert^2 + f(X_s,s)\cp ds + g(X_T)\Big | X_t = x\cb,
\end{equation*}
and the \emph{value function},
$$
V(x,t) = \min_u J(u,x,t).
$$

\paragraph{Verification theorem and Hamilton--Jacobi--Bellman equation.} The verification theorem \citep[see, e.g.,][Chapter VI, Theorem 4.1]{flemingDeterministicStochasticOptimal1975} states that\footnote{We will assume that all functions are regular enough for the verification theorem to apply.} $V$ is characterized by, for all $(x,t) \in \R^d\times [0,T)$, 
$$
\left\{
\begin{array}{l}
     \partial_tV(x,t)  + \min_{u\in \R^d} \ob  L^u_t V(x,t) + \frac{w(t)}{2}\left\Vert u\right\Vert^2 +f(x,t) \cb =0,  \\
     V(x,T) = g(x),
\end{array}
\right.
$$
or equivalently 
$$
\left\{
\begin{array}{l}
     (\partial_t + L_t) V(x,t)   + \min_{u\in \R^d} \ob  \langle \gamma(t)u, \nabla_xV(x,t)\rangle  + \frac{w(t)}{2}\left\Vert u\right\Vert^2 \cb +f(x,t) =0,  \\
     V(x,T) = g(x).
\end{array}
\right.
$$
Solving for the optimal $u$, we get the following equation, often referred to as the Hamilton--Jacobi--Bellman (HJB) equation,
$$
\left\{
\begin{array}{l}
     (\partial_t + L_t) V(x,t) - \frac{1}{2w(t)}\left\Vert (\gamma^\top\nabla V)(x,t)\right\Vert^2 +f(x,t) =0,  \\
     V(x,T) = g(x),
\end{array}
\right.
$$
with the optimal $u$ given by
\begin{equation*}
u^* (x,t) = -\frac{\gamma(t)^\top}{w(t)}\nabla V(x,t).
\end{equation*}
Using the definition of $V$, we get 
\begin{equation*}
\tag{\ref{eq:formula_u_grad_J}}
u^* (x,t) = -\frac{\gamma(t)^\top}{w(t)}\nabla_x J(u^*,x,t).
\end{equation*}
In fact, the equation characterizes the optimal control $u^*$, as stated by the following lemma.
\begin{lemma}
\label{lemma:u_grad_J}
Assume that the control $u$ satisfies 
$$
u (x,t) = -\frac{\gamma(t)^\top}{w(t)}\nabla_x J(u,x,t).
$$
Then $u$ is the optimal control of~\eqref{eq:soc-weighted-gen}.
\end{lemma}

\begin{proof} We prove that if $u$ satisfies the relation, then $J^u: (x,t) \mapsto J(u,x,t)$ solves the HJB equation.
Firstly, we have
\begin{align*}
L_t^u J^u(x,t) &= \lim_{\Delta t \rightarrow 0} \frac{\E[J(u,X^u_{t+\Delta t}, t) |X_t^u = x]  - J(u,x, t)}{\Delta t} \\
&= \lim_{\Delta t \rightarrow 0} \frac{\E[J(u,X^u_{t+\Delta t}, t+\Delta t) |X_t^u = x]  - J(u,x, t)}{\Delta t} +  \E \ob  \frac{ J(u,X^u_{t+\Delta t}, t) - J(u,X^u_{t+\Delta t}, t+\Delta t)}{\Delta t} \Bigg | X_t^u = x\cb \\
&= \lim_{\Delta t \rightarrow 0} \frac{\E[J(u,X^u_{t+\Delta t}, t+\Delta t) |X_t^u = x]  - J(u,x, t)}{\Delta t} - \partial_t J^u(x,t).
\end{align*}
Moreover,
$$
J(u,x,t) = \E \ob \int_t^{t+\Delta t} \frac{w(s)}{2}\Vert u(X_s,s)\Vert^2 + f(X_s, s) ds + J(u,X_{t+\Delta t},t+\Delta t) \Bigg |X_t = x \cb,
$$
hence
$$
\lim_{\Delta t \rightarrow 0} \frac{\E[J(u,X^u_{t+\Delta t}, t+\Delta t) |X_t^u = x]  - J(u,x, t)}{\Delta t} =  - \frac{w(t)}{2}\Vert u(x,t)\Vert^2 - f(x, t).
$$
Therefore, we have
$$
L_t J^u(x,t) = L_t^uJ^u(x,t) - \langle \gamma(t)u(x,t), \nabla_x J^u(x,t) \rangle = - \frac{w(t)}{2}\Vert u(x,t)\Vert^2 - f(x, t)  - \partial_t J^u(x,t) - \langle \gamma(t)u(x,t), \nabla_x J^u(x,t) \rangle.
$$
For $V = J^u$, the HJB equation becomes
\begin{align*}
(\partial_t + L_t)  J^u(x,t) &- \frac{1}{2w(t)}\left\Vert (\gamma^\top\nabla  J^u)(x,t)\right\Vert^2 +f(x,t) =0 \\
&\Leftrightarrow - \frac{w(t)}{2}\Vert u(x,t)\Vert^2  - \langle \gamma(t)u(x,t), \nabla_x J^u(x,t)  \rangle  - \frac{1}{2w(t)}\left\Vert (\gamma^\top\nabla  J^u)(x,t)\right\Vert^2 = 0 \\
&\Leftrightarrow \frac{w(t)}{2}\left\Vert u(x,t) +  \frac{\gamma(t)^\top}{w(t)}\nabla  J^u(x,t) \right\Vert^2 = 0
\end{align*}
which is satisfied under the assumption 
$$
u (x,t) = -\frac{\gamma(t)^\top}{w(t)}\nabla_x J(u,x,t).
$$
By definition, $J^u$ satisfies the boundary condition, as
$$
J^u (x,T) = \E\ob \int_T^T \frac{w(s)}{2}\Vert u(X_s,s)\Vert^2 + f(X_s, s) ds + g(X_T^u) \Big| X_T^u = x \cb = g(x).
$$
By the verification theorem, $J^u = V$, and the optimal control is given by  
$$
u^* (x,t) = -\frac{\gamma(t)^\top}{w(t)}\nabla V(x,t) = -\frac{\gamma(t)^\top}{w(t)}\nabla J^u(x,t) = u (x,t).
$$
\end{proof}

\subsection{The adjoint state}
\paragraph{Adjoint.} For some control $u$, we define the adjoint state as
$$
a(t, \Xb^u, u) = \nabla_{X_t^u}\op \int_t^T\op \frac{w(s)}{2}\Vert u(X_s^u,s)\Vert^2 + f(X_s^u,s)\cp ds + g(X_T^u) \cp
$$
where $\Xb^u$ is the solution to $dX_t^u = \left( b(X_t^u, t) + \gamma(t)\,u(X_t^u, t) \right) dt + \gamma(t)\, dB_t$.
The adjoint state is a random variable that depends on the random trajectory $\Xb^u$, and $\nabla_{X_t^u}$ denotes the derivative with respect to the value $X_t^u$ of the process at time $t$, keeping the trajectory of the Brownian motion $(B_s)_{s\geq t}$ fixed.
For more details on differentiating through solutions of SDEs, see \citet{kidgerNeuralDifferentialEquations,leburuDifferentiatingStochasticDifferential2026}.
We will not discuss technical details in this article and will assume that the processes are regular enough to be differentiated.
Taking the conditional expectation of the adjoint state gives the gradient of the cost function $\nabla J(u,x,t) = \E[a(t,\Xb^u, u)|X_t =x]$.
To compute the adjoint, we will use the fact that it solves an ODE, as stated by the following lemma.
\begin{lemma}
\label{lemma:adjoint_ODE}
The adjoint solves the ODE
\begin{equation}
\label{eq:adjoint-ODE-gen}
\left\{
\begin{array}{cl}
    \frac{d}{dt}a(t, \Xb^u, u) & = - \ob \nabla_{X_t^u}\op b(X_t^u,t) + \gamma(t) u(X_t^u,t)\cp ^\top a(t,\Xb^u,u) + \nabla_{X_t^u}\op f(X_t^u,t) +\frac{w(t)}{2}\Vert u(X_t^u,t)\Vert^2\cp\cb, \\
     a(T, \Xb^u, u)& = \nabla g(X_T^u).
\end{array}
\right.
\end{equation}
\end{lemma}
To simplify notation, we denote  in the rest of this section 
$$b_u: (x,t) \mapsto b(x,t) + \gamma(t) u(x,t),$$
and
$$f_u: (x,t) \mapsto f(x,t) +\frac{w(t)}{2}\Vert u(x,t)\Vert^2.$$ 

\begin{proof}
The adjoint state is 
\begin{align*}
a(t, \Xb^u, u) 
&= \nabla_{X_t^u}\op \int_t^T f_u(X_s^u,s) ds + g(X_T^u) \cp \\
&= \int_t^T  (\nabla_{X_t^u} X_s^u)^\top \nabla_xf_u(X_s^u,s)ds + (\nabla_{X_t^u} X_T^u)^\top \nabla_x g(X_T^u),
\end{align*}
where $\nabla_{X_t^u} X_s^u$ corresponds to the change in $X_s^u$ when changing the value $X_t^u$ at time $t$.
Using the chain rule, for $h > 0$ such that $t+h \leq T$ we have
\begin{align*}
a(t, \Xb^u, u) 
&= \int_t^{t+h}  (\nabla_{X_t^u} X_s^u)^\top \nabla_xf_u(X_s^u,s)ds  + \int_{t+h}^T  (\nabla_{X_t^u} X_s^u)^\top \nabla_xf_u(X_s^u,s)ds+ (\nabla_{X_t^u} X_T^u)^\top \nabla_x g(X_T^u) \\
&= \int_t^{t+h}  (\nabla_{X_t^u} X_s^u)^\top \nabla_xf_u(X_s^u,s)ds\\
&\qquad+ (\nabla_{X_t^u} X_{t+h}^u)^\top \ob\int_{t+h}^T  (\nabla_{X_{t+h}^u} X_s^u)^\top \nabla_xf_u(X_s^u,s)ds+ (\nabla_{X_{t+h}^u} X_T^u)^\top \nabla_x g(X_T^u)\cb\\
&= \int_t^{t+h}  (\nabla_{X_t^u} X_s^u)^\top \nabla_xf_u(X_s^u,s)ds  + (\nabla_{X_t^u} X_{t+h}^u)^\top a(t+h, \Xb^u, u) 
\end{align*}
Hence 
\begin{equation}
\label{lemma:adjoint_ODE:proof:eq:accroiss}
\frac{a(t+h, \Xb^u, u) - a(t, \Xb^u, u)}{h} = -\frac{1}{h}\int_t^{t+h}  (\nabla_{X_t^u} X_s^u)^\top \nabla_xf_u(X_s^u,s)ds + \op\frac{I - \nabla_{X_t^u} X_{t+h}^u}{h}\cp^\top a(t+h, \Xb^u, u) 
\end{equation}
As $\Xb^u$ is the solution to the SDE $dX_t^u = b_u(X_t^u,t)dt + \gamma(t)dB_t$,
$$
X_{t+h}^u = X_t^u + \int_t^{t+h}b_u(X_s^u,s)ds + \int_t^{t+h}\gamma(s)dB_s,
$$
hence 
$$
\nabla_{X_t^u}X_{t+h}^u = I +  \int_t^{t+h} \nabla_xb_u(X_s^u,s)(\nabla_{X_t^u}X_s^u)ds,
$$
leading to 
$$
\lim_{h\rightarrow 0 }\frac{I - \nabla_{X_t^u} X_{t+h}^u}{h} = -\nabla_xb_u(X_t^u,t)(\nabla_{X_t^u}X_t^u) =  -\nabla_xb_u(X_t^u,t).
$$
Taking $h\rightarrow0$ in~\eqref{lemma:adjoint_ODE:proof:eq:accroiss} gives
$$
\frac{d}{dt}a(t, \Xb^u, u) = - (\nabla_{X_t^u} X_t^u)^\top \nabla_xf_u(X_t^u,t) -\nabla_xb_u(X_t^u,t)^\top a(t, \Xb^u, u) = - \nabla_xf_u(X_t^u,t) -\nabla_xb_u(X_t^u,t)^\top a(t, \Xb^u, u)
$$
Finally,
$$
a(T, \Xb^u, u) = \nabla_{X_T^u}\op \int_T^T f_u(X_s^u,s) ds + g(X_T^u) \cp = \nabla g(X_T^u).
$$
\end{proof}

Integrating~\eqref{eq:adjoint-ODE-gen} gives 
$$
a(t, \Xb^u, u) = \nabla g(X_T^u) + \int_t^T\op \nabla_{x}b_u(X_s^u,s)^\top a(s,\Xb^u,u) + \nabla_{x}f_u(X_s^u,s)\cp ds
$$
Denoting $Y_t = \nabla_xJ(u,X_t,t) = \E[a(t, \Xb^u, u)|X_t]$ and taking the conditional expectation in the equation above, we have
\begin{align}
Y_t 
&= \E\ob \nabla g(X_T^u) + \int_t^T\op \nabla_xb_u(X_s^u,s)^\top a(s,\Xb^u,u) + \nabla_xf_u(X_s^u,s)\cp ds \Bigg| X_t\cb \notag\\
&= \E\ob \nabla g(X_T^u) + \int_t^T\op \nabla_xb_u(X_s^u,s)^\top \E[a(s,\Xb^u,u)|X_s] + \nabla_xf_u(X_s^u,s)\cp ds \Bigg| X_t\cb \notag\\
&= \E\ob \nabla g(X_T^u) + \int_t^T\op \nabla_xb_u(X_s^u,s)^\top Y_s + \nabla_xf_u(X_s^u,s)\cp ds \Bigg| X_t\cb. \label{eq:adjoint_cond_exp}
\end{align}
We prove in the following lemma that equation~\eqref{eq:adjoint_cond_exp} characterizes $Y_t$.
\begin{lemma}
\label{lemma:unicity_adjoint}
Let $A : \R^d\times[0,T] \rightarrow \R^{d\times d}$ and $c : \R^d\times[0,T] \rightarrow \R^d$ be measurable.
Assume that there is a measurable $\kappa : (0,T] \rightarrow \R_+$ such that $\Vert A(x,s)\Vert \leq \kappa(s)$ in operator norm for all $(x,s) \in \R^d\times(0,T]$, and $\int_\ep^T \kappa(s) ds < \infty$ for every $\ep \in (0,T]$.
Suppose that $Y^1$ and $Y^2$ are two processes of the form $Y^i_t = y^i(X_t^u,t)$, with $y^i$ measurable and $\sup_{t\in[0,T]}\E\Vert Y^i_t\Vert < \infty$, that both satisfy, for all $t \in [0,T]$,
$$
Y_t = \E\ob \nabla g(X_T^u) + \int_t^T\op A(X_s^u,s)^\top Y_s + c(X_s^u,s)\cp ds \Bigg| X_t^u\cb ,
$$
where $\nabla g(X_T^u)$, $\int_0^T \Vert c(X_s^u,s)\Vert ds$ and $\int_0^T \Vert A(X_s^u,s)^\top Y^i_s\Vert ds$ are assumed integrable.
Then $Y^1 = Y^2$.
\end{lemma}

\begin{proof}
Let $D_t = Y_t^1 - Y_t^2$ and $\varphi(t) = \E\Vert D_t\Vert$. Subtracting the two equations, the terminal and source terms cancel
$$
D_t = \E\ob\int_t^T A(X_s^u,s)^\top D_s ds \Bigg| X_t^u\cb .
$$
Using the triangle inequality and the bound on $A$, we get
$$
\Vert D_t \Vert \leq \E\ob\int_t^T \Vert A(X_s^u,s)^\top D_s\Vert ds \Bigg| X_t^u\cb \leq \E\ob\int_t^T \kappa(s)\Vert D_s\Vert ds \Bigg| X_t^u\cb ,
$$
and taking expectations leads to
$$
\varphi(t) \leq \int_t^T \kappa(s)\varphi(s) ds .
$$
Fix $\ep \in (0,T]$.
As $\kappa$ is integrable on $[\ep,T]$, Grönwall's lemma gives $\varphi = 0$ on $[\ep,T]$.
As $\ep$ is arbitrary, $D_t = 0$ almost surely for every $t \in (0,T]$.
For $t=0$, we then have
$$
\E \Vert D_0 \Vert \leq \E\ob\int_0^T \Vert A(X_s^u,s)^\top D_s\Vert ds \cb = \int_0^T \E\Vert A(X_s^u,s)^\top D_s\Vert ds = \int_0^T0ds = 0,
$$
hence $D_0 = 0$ almost surely.
\end{proof}

\paragraph{Remark: the assumption on $A$ holds for the flow matching drift.}
In the flow matching setting of the main text, $T=1$ and we have
$$
\nabla_x b(x,t) = 2\nabla_x v^\base(x,t) - \frac{\dot\alpha_t}{\alpha_t} I,
\qquad
\nabla_x b_{u_\theta}(x,t) = 2\nabla_x v^\ft_\theta(x,t) - \frac{\dot\alpha_t}{\alpha_t} I.
$$
If $v^\base$ and $v^\ft_\theta$ are $L$-Lipschitz in $x$ uniformly in $t$, then
$$
\Vert \nabla_x b(x,t)\Vert \leq 2L + \frac{\dot\alpha_t}{\alpha_t} =: \kappa(t),
\qquad
\int_\ep^1 \kappa(t) dt = 2L(1-\ep) + \log\frac{1}{\alpha_\ep} < \infty
$$
for every $\ep \in (0,1]$, and the same bound holds for $\nabla_x b_{u_\theta}$.
For $\alpha_t = t$, we have $\kappa(t) = 2L + 1/t$, which is not an issue: the lemma only requires integrability away from the initial time, because the backward Grönwall argument starts at the terminal time $t = 1$.
For a general control $u$ Lipschitz in $x$, $\nabla_x b_u = \nabla_x b + \gamma(t)\nabla_x u$ adds the term $\gamma(t) L_u$, which is bounded on $[\ep,1]$.

\subsection{Proof of Proposition~\ref{prop:critical_point}}
We start by restating Proposition~\ref{prop:critical_point} more precisely:
\begin{proposition}
\label{prop:critical_point-gen}
For the loss defined by
\begin{align}
\label{eq:adjoint-matching-weighted-gen}
\hat \L_{\textnormal{WAM}}(u,v) &:= \E_{\Xb \sim p^{v}}\ob\frac{1}{2}\int_0^T\left\Vert u(X_t, t) + \frac{\gamma(t)^\top \td a (t,\Xb)}{w(t)} \right\Vert^2 dt\cb,  \\
\textnormal{where}\qquad &\frac{d}{dt} \td a(t, \Xb)  = - (\nabla_xb(X_t,t)^\top\td a(t,\Xb) + \nabla_x f(X_t, t)),\notag\\
& \td a(T,\Xb) = \nabla_x g(X_T),\notag
\end{align}
the only critical point $\frac{\delta}{\delta u}\hat\L_{\textnormal{WAM}}(u,v)|_{v=u} = 0$ is $u^*$, the optimal control of the weighted SOC problem~\eqref{eq:soc-weighted-gen}.
\end{proposition}
The rest of this section is dedicated to the proof of Proposition~\ref{prop:critical_point-gen}.

\paragraph{Functional gradient.}
For a measure $\M$ on $\R^d\times[0,T]$, a functional $\F: L^2(\R^d\times[0,T],\M;\R^d) \rightarrow \R$ and $f \in  L^2(\R^d\times[0,T],\M;\R^d)$, we define the functional gradient $\frac{\delta}{\delta f}\F(f)$ as the function $g \in  L^2(\R^d\times[0,T],\M;\R^d)$ such that for all $h \in  L^2(\R^d\times[0,T],\M;\R^d)$
$$
\lim_{\ep \rightarrow 0} \frac{\F(f+\ep h)-\F(f)}{\ep} = \int \langle g(x,t), h(x,t) \rangle \M(dx,dt).
$$

\paragraph{First step, critical point condition.} For $v$ a fixed control, and the measure $\M(dx,dt) = p(X_t^v = x)dxdt$ we have 
\begin{align*}
\E_{\Xb \sim p^v}[&\L_{\textnormal{WAM}}(u,\Xb)] \\
&= \E_{\Xb \sim p^v}\ob\frac{1}{2}\int_0^T\left\Vert u(X_t, t) + \frac{\gamma(t)^\top \td a (t,\Xb)}{w(t)} \right\Vert^2 dt\cb\\
&= \frac{1}{2}\int_0^T\E_{\Xb \sim p^v}\ob\Vert u(X_t, t)\Vert^2\cb dt + \int_0^T\E_{\Xb \sim p^v}\ob \left\langle u(X_t, t),\frac{\gamma(t)^\top \td a (t,\Xb)}{w(t)} \right\rangle\cb dt + \textnormal{cst} \\
&= \frac{1}{2}\int_0^T\E_{\Xb \sim p^v}\ob\Vert u(X_t, t)\Vert^2\cb dt + \int_0^T\E_{\Xb \sim p^v}\ob \left\langle u(X_t, t),\frac{\gamma(t)^\top \E_{\Xb \sim p^v}[\td a (t,\Xb)|X_t]}{w(t)} \right\rangle\cb dt + \textnormal{cst} \\
&=\int \op\frac{1}{2}\Vert u(x,t)\Vert^2 + \left\langle u(x, t),\frac{\gamma(t)^\top \E_{\Xb \sim p^v}[\td a (t,\Xb)|X_t=x]}{w(t)} \right\rangle\cp \M(dx,dt).
\end{align*}
It follows that, for $(x,t)\in \R^d\times[0,T]$,
$$
\frac{\delta}{\delta u} \E_{\Xb \sim p^v}[\L_{\textnormal{WAM}}(u,\Xb)](x,t)= u(x,t)+ \frac{\gamma(t)^\top \E_{\Xb \sim p^v}[\td a (t,\Xb)|X_t=x]}{w(t)},
$$
hence $u$ is a critical point of  $u\mapsto \E_{\Xb \sim p^{\bar u}}[\L_{\textnormal{WAM}}(u,\Xb)]$ if and only if
\begin{equation}
\label{proof:prop:critical_point:crit_eq}
\forall (x,t)\in \R^d\times[0,T], \quad u (x,t)  = -\frac{\gamma(t)^\top}{w(t)}\E[\td a(t,\Xb^{u})|X_t^u =x]   
\end{equation}

\paragraph{Second step, lean adjoint equation.} By construction $\td a(t,\Xb^u)$ satisfies for $t\in [0,T]$
$$
\td a(t,\Xb^u) = \nabla_xg(X_T^u) + \int_t^T (\nabla_xb(X_s^u,s)^\top \td a(s,\Xb^u) +\nabla_x f(X_s^u,s))ds.
$$
Denoting $\td Y_t = \E[\td a(t,\Xb^u)|X_t]$, we get that
\begin{equation}
\label{eq:lean_adjoint_cond_exp}
\td Y_t = \E\ob \nabla_xg(X_T^u) + \int_t^T (\nabla_xb(X_s^u,s)^\top \td Y_s +\nabla_x f(X_s^u,s))ds \Big|X_t\cb.
\end{equation}

\paragraph{Third step, critical point implies optimal control.}
Multiplying~\eqref{proof:prop:critical_point:crit_eq} by $\nabla_x u(x,t)^\top$, we get
\begin{multline*}
\nabla_x u (x,t)^\top  u (x,t)= \nabla_x u (x,t)^\top\op-\frac{\gamma(t)^\top}{w(t)}\E[\td a(t,\Xb^{u})|X_t^{u} =x]\cp \\
\Leftrightarrow \E\ob w(t)\nabla_x u (X_t^{u},t)^\top u (X_t^{u},t) + \nabla_{x}\op\gamma(t) u (X_t^{u},t)\cp^\top\td a(t,\Xb^{u}) \Big|X_t =x\cb =0.
\end{multline*}
Plugging this equality into~\eqref{eq:lean_adjoint_cond_exp} gives
$$
\td Y_t = \E\ob \nabla_xg(X_T^u) + \int_t^T (\nabla_xb_u(X_s^u,s)^\top \td Y_s +\nabla_x f_u(X_s^u,s))ds \Big|X_t\cb.
$$
By Lemma~\ref{lemma:unicity_adjoint}, applied with $A = \nabla_x b_u$ and $c = \nabla_x f_u$, $\td Y_t = Y_t = \E[a(t, \Xb^u, u)|X_t]$, i.e., 
$$
u(x,t) = -\frac{\gamma(t)^\top}{w(t)}\E[a(t,\Xb^{u},u)|X_t^u =x] = -\frac{\gamma(t)^\top}{w(t)}\nabla_x J(u,x,t), 
$$
and Lemma~\ref{lemma:u_grad_J} gives $u = u^*$: a critical point is necessarily the optimal control.

\paragraph{Fourth step, optimal control is a critical point.} Conversely, we check that $u^*$ satisfies~\eqref{proof:prop:critical_point:crit_eq}, i.e., that it is a critical point.
Take $u = u^*$.
By~\eqref{eq:formula_u_grad_J}, $Y_t = \E[a(t,\Xb^{u^*},u^*)|X_t^{u^*}] = \nabla_x J(u^*,X_t^{u^*},t)$ satisfies $u^*(X_t^{u^*},t) = -\frac{\gamma(t)^\top}{w(t)}Y_t$, hence for all $s\in[0,T]$,
\begin{align*}
\nabla_x b_{u^*}(X_s^{u^*},s)^\top Y_s + \nabla_x f_{u^*}(X_s^{u^*},s)
&= \nabla_x b(X_s^{u^*},s)^\top Y_s + \nabla_x f(X_s^{u^*},s) + \nabla_x u^*(X_s^{u^*},s)^\top\op\gamma(s)^\top Y_s + w(s)u^*(X_s^{u^*},s)\cp\\
&= \nabla_x b(X_s^{u^*},s)^\top Y_s + \nabla_x f(X_s^{u^*},s).
\end{align*}
Plugging this identity into the equation~\eqref{eq:adjoint_cond_exp} satisfied by $Y_t$, we get
$$
Y_t = \E\ob \nabla_xg(X_T^{u^*}) + \int_t^T (\nabla_xb(X_s^{u^*},s)^\top Y_s +\nabla_x f(X_s^{u^*},s))ds \Big|X_t^{u^*}\cb,
$$
which is also the equation~\eqref{eq:lean_adjoint_cond_exp} satisfied by $\td Y_t$.
Lemma~\ref{lemma:unicity_adjoint}, applied with $A = \nabla_x b$ and $c = \nabla_x f$, gives $Y = \td Y$, hence
$$
u^*(x,t) = -\frac{\gamma(t)^\top}{w(t)}\E[\td a(t,\Xb^{u^*})|X_t^{u^*} = x],
$$
which is~\eqref{proof:prop:critical_point:crit_eq} for $u = u^*$.

\paragraph{Adding an additional time-dependent weight.} Adding an additional time-dependent weight $\lambda(t) > 0$, as in~\eqref{eq:adjoint-matching-weighted-stab}
$$
\L_{\textnormal{WAM}}(u,\Xb) := \frac{1}{2}\int_0^T\lambda(t)\left\Vert u(X_t, t) + \frac{\gamma(t)^\top \td a (t,\Xb)}{w(t)} \right\Vert^2  dt
$$
does not change the critical points, as it only changes the base distribution $\M(dx,dt) = \lambda(t) p(X_t^v = x)dxdt$ to compute the functional gradient.

\subsection{Link with \citet{domingo-enrichAdjointMatchingLens2026}}
\label{appendix:weighted_adjoint_matching_algo:connection}
Recently, \citet{domingo-enrichAdjointMatchingLens2026} proposed a more general Hamiltonian framework for adjoint matching.
They derived the corresponding adjoint matching loss for a general problem of the form 
\begin{align*}
\min_{u \in \mathcal{U}} \quad & \mathbb{E} \ob \int_0^T  f(X_t^u,u(X_t^u,t),  t)  dt + g(X_T^u) \cb \\
\text{s.t.} \quad & dX_t^u = b(X_t^u,u(X_t^u,t), t)  dt + \gamma(X_t^u,u(X_t^u,t), t) \, dB_t, \\
& X_0^u \sim p_0.
\end{align*}
Taking 
$$
b(x,u,t) = b(x,t) + \gamma(t)u, \quad f(x,u,t) = f(x,t) + \frac{1}{2}\Vert u\Vert^2, \quad \gamma(x,u,t) = \gamma(t),
$$
they recover exactly  (Theorem 2 and Remark 3) the adjoint matching from \citet{domingo-enrichAdjointMatchingFinetuning2025}.
We can also apply their result (Theorem 2) to our weighted SOC problem~\eqref{eq:soc-weighted-gen}, with 
$$
b(x,u,t) = b(x,t) + \gamma(t)u, \quad f(x,u,t) = f(x,t) + \frac{w(t)}{2}\Vert u\Vert^2, \quad \gamma(x,u,t) = \gamma(t),
$$
leading to an adjoint matching formulation
\begin{align*}
\L_{\textnormal{WAM}}(u,\Xb) &:= \frac{1}{2}\int_0^Tw(t)\left\Vert u(X_t, t) + \frac{\gamma(t)^\top \td a (t,\Xb)}{w(t)} \right\Vert^2  dt, \quad \Xb \sim p^{\bar u}, \quad \bar u = \st(u) \\
\textnormal{where}\qquad &\frac{d}{dt} \td a(t, \Xb)  = - (\nabla_xb(X_t,t)^\top\td a(t,\Xb) + \nabla_x f(X_t, t)),\\
& \td a(T,\Xb) = \nabla_x g(X_T).
\end{align*}
This is~\eqref{eq:adjoint-matching-weighted-stab} with $\lambda(t) = w(t)$.
The framework of \citet{domingo-enrichAdjointMatchingLens2026} therefore recovers our loss for one choice of $\lambda$, but lacks the extra degree of freedom that $\lambda$ provides.

\subsection{Description of the algorithm}
\label{appendix:alg:weighted-adjoint-matching}

To run the algorithm, we first discretize the generative SDE~\eqref{eq:soc-weighted:SDE}, giving~\eqref{eq:alg:discreteSDE}, then the adjoint ODE~\eqref{eq:adjoint-matching-weighted:adjoint_ODE}, leading to~\eqref{eq:alg:discreteadjointODE}.
We get a stochastic estimate of the loss $\hat{\mathcal{L}}_{\textnormal{WAM}}(\theta)$ by averaging over trajectories and timesteps.
We then take a stochastic gradient descent step on $\theta$.
The full procedure is summarized in Algorithm~\ref{alg:weighted-adjoint-matching}.
In our implementation, we also subsample $K_0 < K$ timesteps at random in~\eqref{eq:alg:objective} to reduce its computational cost.
For the classical choice $\alpha_t = t$, $\beta_t = 1-t$, two terms of~\eqref{eq:alg:discreteSDE}, $\frac{\dot\alpha_{t_k}}{\alpha_{t_k}} = \frac{1}{t_k}$ and $\gamma(t_k) = \sqrt{\frac{2(1-t_k)}{t_k}}$, are singular at $t_0=0$.
For the first step only, we evaluate these terms at $t_1 = h > 0$ instead of $t_0$.

\begin{algorithm*}[t]
\caption{Weighted Adjoint Matching to finetune a flow matching model}
\begin{algorithmic}
\State \Input Flow matching model $v^\textnormal{base}=v^\textnormal{base}_{\theta_0}, \alpha_t,\beta_t$, reward function $r$, control weight $w(t)$, loss weight $\lambda(t)$,  number of discretization steps $K$, number of trajectories per finetuning iteration $M$, number of finetuning iterations $N$.
\State Initialize parameters of the finetuned model $\theta=\theta_0$, i.e., $v^\ft_\theta = v^\textnormal{base}$. Define $h = \frac{1}{K}$ and for~${k \in \{0,1,\dots, K\}}, t_k = kh$.
\For{$n \in \{0, \ldots, N-1\}$}
    \State Sample $M$ trajectories $\Xb^{(m)} = (X^{(m)}_{t_k})_{k \in \{0,\dots,K\}}$ for $m \in \{ 0,..., M-1\}$
    \begin{equation}
    \label{eq:alg:discreteSDE}
    \left\{
    \begin{array}{ll}
    X^{(m)}_{t_0} & = X^{(m)}_0 \sim  \N(0,I), \\
    X^{(m)}_{t_{k+1}} & = X^{(m)}_{t_k} + h\op2v^{\ft}_\theta(X^{(m)}_{t_k}, t_k) - \frac{\dot\alpha_{t_k}}{\alpha_{t_k}}X^{(m)}_{t_k}\cp + \sqrt{h} \gamma(t_k) \N(0,I).
    \end{array}
    \right.
    \end{equation}
    \State Solve the adjoint ODE backwards in time $\td{\mathbf{a}}^{(m)} = (\td{a}^{(m)}_{t_k})_{k \in \{0,\dots,K\}}$
    \begin{equation}
    \label{eq:alg:discreteadjointODE}
    \left\{
    \begin{array}{ll}
    \td{a}^{(m)}_{t_K}&=  \td{a}^{(m)}_1 = -a \nabla r(X^{(m)}_1), \\
    \td{a}^{(m)}_{t_{k-1}} &= \td{a}^{(m)}_{t_k} + h\nabla_{X^{(m)}_{t_k}}\left(2v^{\textnormal{base}}(X^{(m)}_{t_k}, t_k) - \frac{\dot\alpha_{t_k}}{\alpha_{t_k}}X^{(m)}_{t_k}\right)^\top \td{a}^{(m)}_{t_k} .
    \end{array}
    \right.
    \end{equation}
    \State Take $\Xb^{(m)} = \st(\Xb^{(m)})$ and $\td{\mathbf{a}} = \st(\td{\mathbf{a}})$.
    \State Compute the Weighted Adjoint Matching objective:
    \begin{equation}
    \label{eq:alg:objective}
        \hat{\mathcal{L}}_{\textnormal{WAM}}(\theta) = \frac{1}{2M(K+1)}\sum_{m,k} \lambda(t_k) \left\Vert \frac{2}{\gamma(t_k)} \left(v^{\ft}_\theta(X^{(m)}_{t_k}, t_k) - v^{\textnormal{base}}(X^{(m)}_{t_k}, t_k)\right) + \frac{\gamma(t_k)}{w(t_k)}\td{a}^{(m)}_{t_k} \right\Vert^2. 
    \end{equation}
    \State Compute the gradient $\nabla_\theta \hat{\mathcal{L}}_{\textnormal{WAM}}(\theta)$ and update the parameters $\theta$ using gradient descent.
\EndFor
\State \Output Finetuned model $v^{\ft}_\theta$
\end{algorithmic}
\label{alg:weighted-adjoint-matching}
\end{algorithm*}

\section{Experiment details and additional experiments}
\label{appendix:additional_experiments}

\subsection{Experiment details}

\paragraph{Compute infrastructure.}
The experiments were run on an academic Slurm cluster. Each job was launched on a single NVIDIA GPU (RTX 2080Ti/6000/8000, V100, A100, H100 or H200), with 4--8 CPU cores and up to 12~GB of RAM.
The experiments, including preliminary experiments, experiments not reported in the paper and jobs that were preempted or lost, ran for about 480 GPU-hours.
The source code will be released at the time of publication.

\paragraph{Gaussian mixtures.}
For the first Gaussian mixture experiment of Section~\ref{sct:exp:GM}, the base model is an MLP with $4\,546$ parameters, trained using Adam (constant
learning rate $2\times10^{-4}$) and a batch size of $100$ for $100$ epochs, each epoch using $10^5$ fresh samples of the target.
We use the exponential moving average (with decay $0.999$) of its weights.
Finetuning is done over 80 epochs of $10$ batches of $1000$ trajectories, sampled using $40$ discretization steps.
The optimizer is Adam (learning rate $2\times10^{-4}$), and we use an exponential moving average of the weights with decay $0.99$.
We sweep $a\in[0.2, 39.95]$ for adjoint matching and for the piecewise constant ($t_s\in\{0.25, 0.5, 0.65, 0.8\}$), linear ($w_0 \in\{10, 10^2, 10^3, 10^4\}$) and exponential ($\alpha\in\{2, 4, 6, 8\}$) weighting functions.
For each method (adjoint, piecewise, exponential or linear with fixed value of $t_s$, $\alpha$ or $w_0$), we start from the smallest value of $a$, then warm start the model's weights using weights from the previous value of $a$, and run the optimization over only $40$ epochs.
The final checkpoints are evaluated on $10^6$ samples.

For the additional experiment of Appendix~\ref{appendix:additional_experiments:exp-GM-shift}, the experiment parameters are similar except for $a\in[0.2, 29.95]$ and we did not use the linear weighting function.
For the additional experiment of Appendix~\ref{appendix:additional_experiments:exp-GM-four}, the experiment parameters are similar except for $a\in[0.2, 9.2]$ and $t_s\in\{0.5, 0.6, 0.7, 0.8\}$, no warm start, and we did not use the linear and exponential weighting functions.

\paragraph{MNIST.}
For the MNIST experiments of Section~\ref{sct:exp:MNIST} and Appendix~\ref{appendix:additional_experiments:exp-MNIST-dark}, the base model is a UNet with $9.0$ million parameters, trained on the MNIST
training images rescaled to $[-1, 1]$, with batch size $64$, using Adam (learning rate $2\times10^{-4}$,
halved by PyTorch's \texttt{ReduceLROnPlateau} on the validation loss, patience 5) for $50$ epochs.
We use the exponential moving average (with decay $0.9999$) of its weights.
Each finetuning run starts from the base model and lasts $25$ epochs of $50$ batches of $64$
trajectories, sampled with $40$ discretization steps.
We use Adam (learning rate
$2\times10^{-4}$, halved by \texttt{ReduceLROnPlateau} on the mean gradient norm of each epoch,
patience 5) and an exponential moving average of the weights with decay $0.995$.
We sweep $a\in[1, 10]$ for adjoint matching and, for the piecewise constant weighting function, $a\in[10, 50]$ at
$t_s=0.25$, $[10, 100]$ at $t_s=0.3$, $[40, 150]$ at $t_s=0.35$, $[80, 250]$ at $t_s=0.4$ and
$[120, 375]$ at $t_s=0.45$.
The final checkpoints are evaluated on $5\times10^4$ samples and classified by the Hugging Face classifier
\texttt{farleyknight/mnist-digit-classification-2022-09-04} (a finetuned ViT-B/16, Apache license 2.0).

\subsection{Gaussian mixtures}
\subsubsection{Using a linear weighting function}
\begin{figure*}
    \centering
    \includegraphics[width=.55\linewidth]{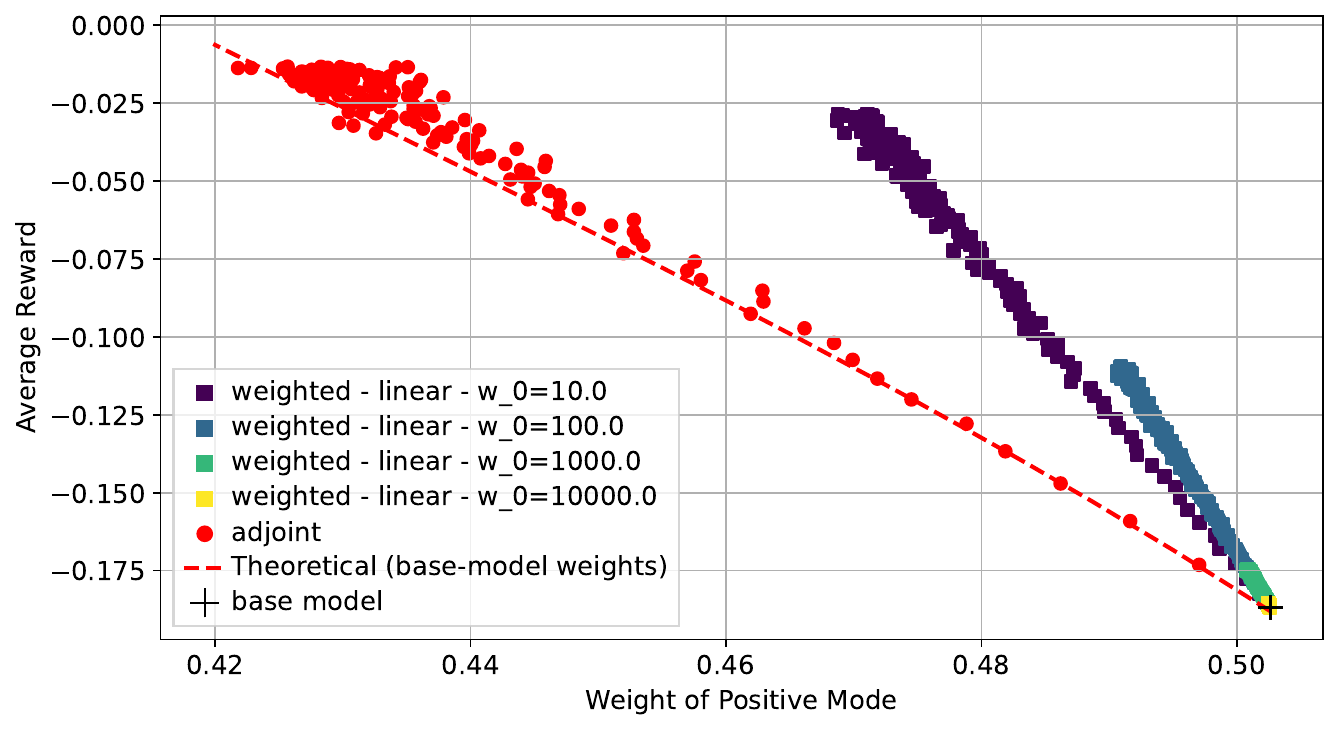}
    \caption{Plot of $\E[r(\hat X)]$ with respect to $\E[\1_{x_1>0}(\hat X)]$ for adjoint matching and weighted adjoint matching with a linear weighting function.}
	\label{fig:append:reward_weight_GM:linear}
\end{figure*}

For the experiment of Section~\ref{sct:exp:GM}, we also tried using a linear function $w_\textrm{linear}(t) = w_0 + (w_1-w_0)t$, with $w_1=1$ and different values for $w_0$. In Figure~\ref{fig:append:reward_weight_GM:linear}, we observe that the linear weighting function achieves a better balance/reward tradeoff than classical adjoint matching, but is less effective than the exponential or piecewise constant weighting function.

\subsubsection{Gaussian mixture: a shifted mode}
\label{appendix:additional_experiments:exp-GM-shift}
\begin{figure}
    \centering
    \captionsetup[subfigure]{justification=centering}
    \subfloat[][Adjoint matching and weighted adjoint matching with an exponential weighting function.]{
    \includegraphics[width=.55\linewidth]{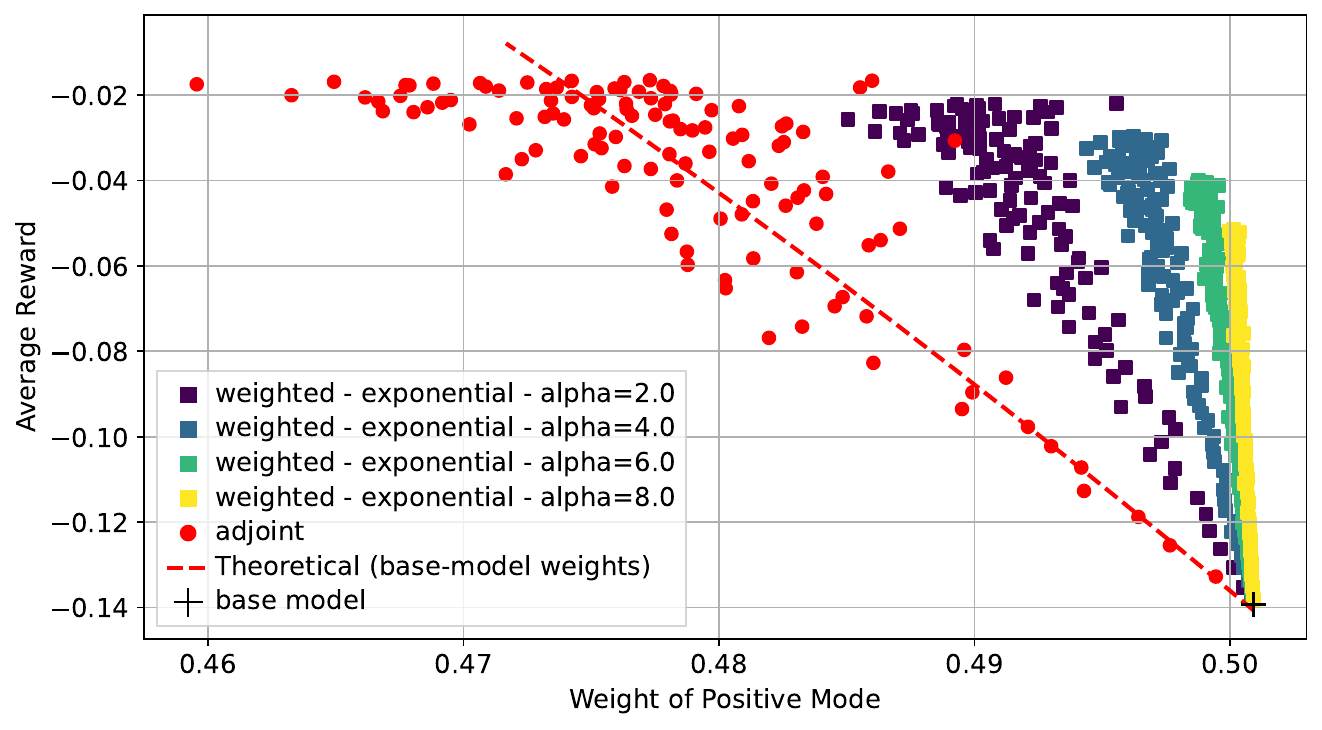}
    }\\
    \subfloat[][Adjoint matching and weighted adjoint matching with a piecewise constant weighting function.]{
    \includegraphics[width=.55\linewidth]{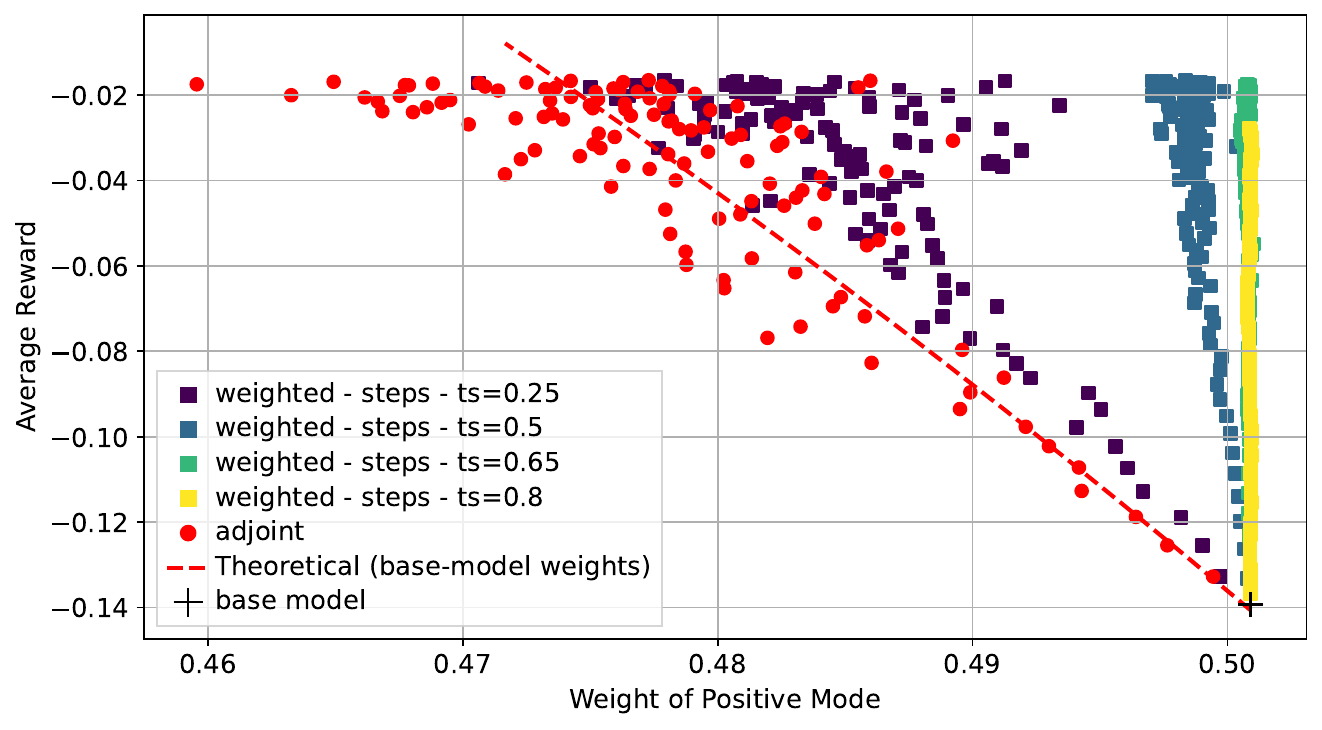}
    }
    \caption{Average reward with respect to weight of positive mode for adjoint matching and weighted adjoint matching, using different functions $t\mapsto w(t)$.}
	\label{fig:reward_weight_GM-shift}
\end{figure}

In Figure~\ref{fig:reward_weight_GM-shift}, we reproduce the experiment of Section~\ref{sct:exp:GM} with a different base distribution (two modes with the same variance but the second one is shifted off the first axis)
$$
p_\target(x) = \frac{1}{2}\N(x;\mu_1,\sigma^2I) + \frac{1}{2}\N(x;\mu_2,\sigma^2I) = \frac{1}{2} (2 \pi \sigma^2)^{-1}e^{-\frac{\Vert x - \mu_1 \Vert^2}{2 \sigma^2}} + \frac{1}{2} (2 \pi \sigma^2)^{-1}e^{-\frac{\Vert x - \mu_2 \Vert^2}{2 \sigma^2}},
$$
with $\mu_1 = (-m,0)$, $\mu_2 = (m,s)$, $m,s\in\R$ and $\sigma >0$.
The conclusions are unchanged.

\subsubsection{Gaussian mixture: four modes}
\label{appendix:additional_experiments:exp-GM-four}
\begin{figure}[p]
    \centering
    \includegraphics[width=0.9\linewidth]{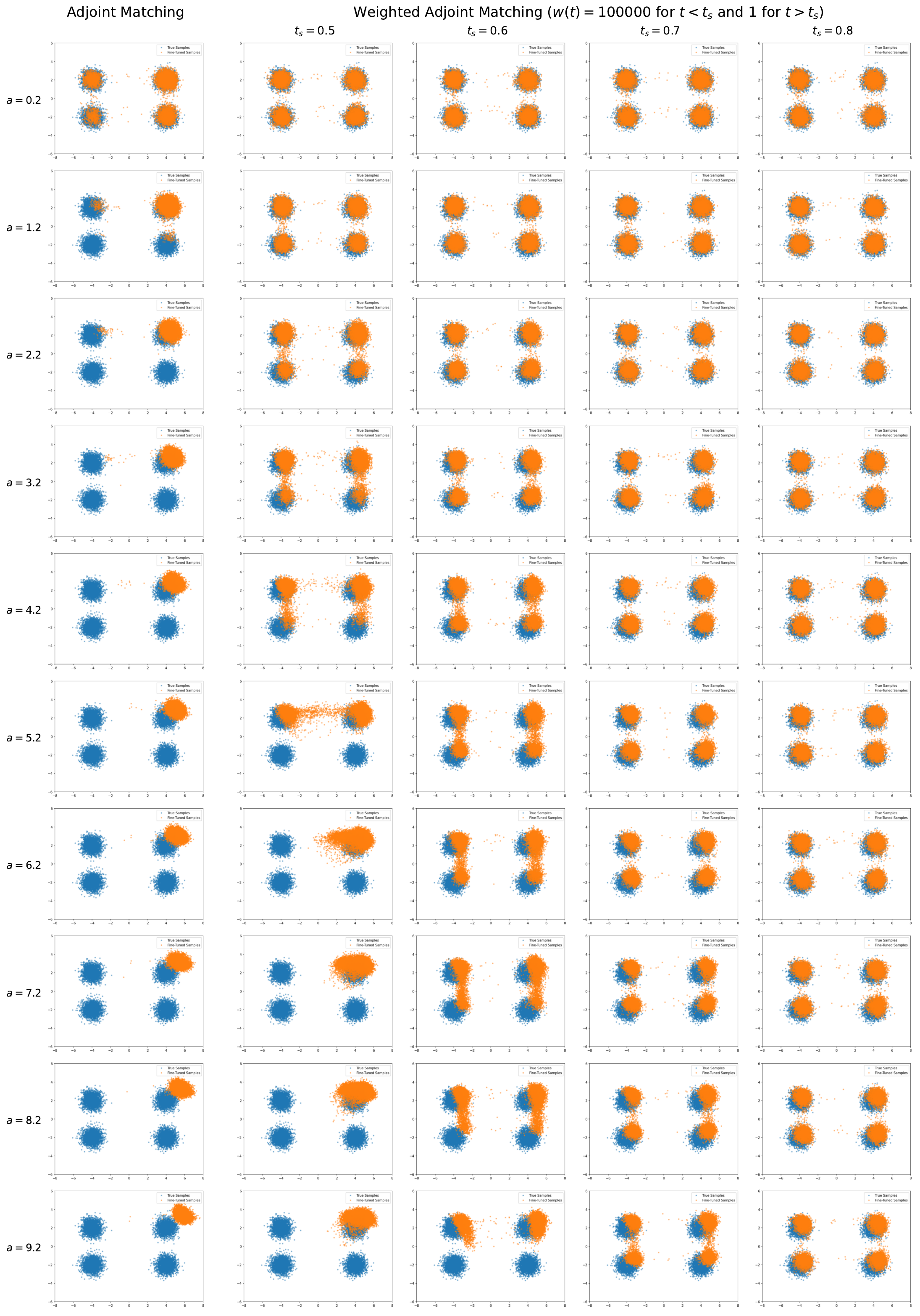}
    \caption{Sample of base distribution (blue) and finetuned models (orange), for adjoint matching and weighted adjoint matching for different values of $a$ and $t_s$.}
	\label{fig:reward_weight_GM-four}
\end{figure}

We ran an additional experiment with a mixture of 4 Gaussians as base distribution
$$
p_\target(x) = \frac{1}{4}\sum_{i=1}^4\N(x;\mu_i,\sigma^2I),
$$
with $\mu_1 = (4, -2), \mu_2 = (4, 2), \mu_3 = (-4, -2), \mu_4  = (-4, 2)$ and $\sigma^2 =0.25$.
The reward is $r(x) = x_1 + x_2$ and grows by moving to the upper right.
In Figure~\ref{fig:reward_weight_GM-four}, we plot samples of the base distribution alongside samples of the output of adjoint matching and weighted adjoint matching with the piecewise constant function $w_\textrm{steps}(t)$ for different values of $a$ and $t_s$.
While adjoint matching quickly collapses to the top-right mode, we see that this effect is mitigated by weighted adjoint matching depending on the value of $t_s$.
Collapse is still observed with $t_s=0.5$, though it happens later.
For $t_s=0.6$, we have a collapse within each pair of vertically adjacent modes, but not a collapse between left and right.
This is because the horizontal spacing of the modes is larger than the vertical spacing, hence the left-right splitting happens earlier along the generative trajectory. Here the left-right feature is the coarse feature that is preserved, while the top-bottom feature is the finer feature that is changed during finetuning.
Finally, for $t_s =0.7$ and $t_s =0.8$, we observed that both left-right and top-bottom features are preserved, and reward maximization happens only within each mode by shifting locally the samples to the upper right.

\subsection{MNIST main experiment}
\begin{figure}[p]
    \centering
    \captionsetup[subfigure]{justification=centering}
    \subfloat[][Base model, average reward $\hat R = 0.140$]{
    \includegraphics[width=.75\linewidth]{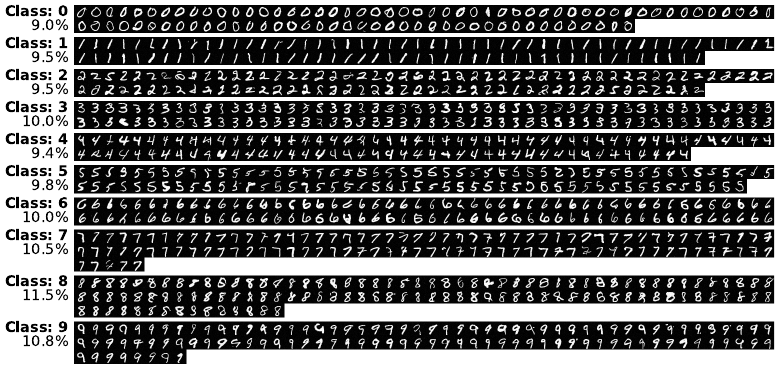}
    }\\
    \subfloat[][Adjoint matching, $a=5.7$, $\hat R = 0.159$]{
    \includegraphics[width=.75\linewidth]{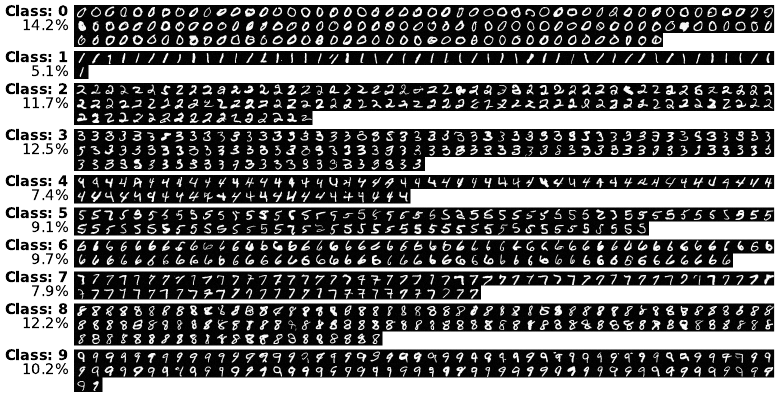}
    }\\
    \subfloat[][Weighted adjoint matching, $t_s = 0.45$, $a=120$, $\hat R = 0.160$]{
    \includegraphics[width=.75\linewidth]{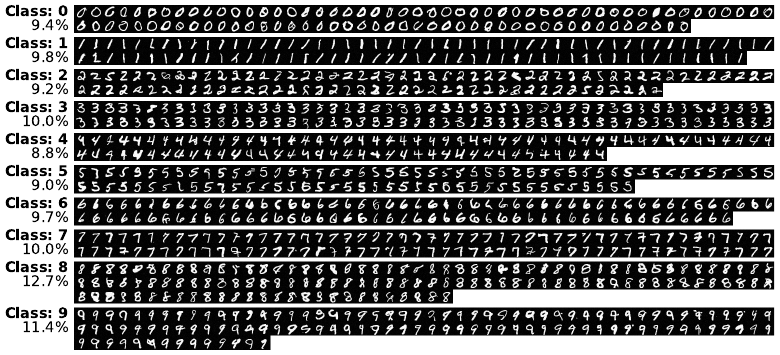}
    }
    \caption{Generated samples grouped by predicted class, for the run of each method whose average luminosity is closest to 0.16.
    Percentages are computed on the $1\,000$ displayed samples, and therefore differ from those of Table~\ref{tab:exp-MNIST_class_percentages_0.16}, computed on $50\,000$ samples.}
    \label{fig:exp-MNIST_class_balance_0.16}
\end{figure}

\begin{figure}[p]
    \centering
    \captionsetup[subfigure]{justification=centering}
    \subfloat[][Base model, average reward $\hat R = 0.140$]{
    \includegraphics[width=.75\linewidth]{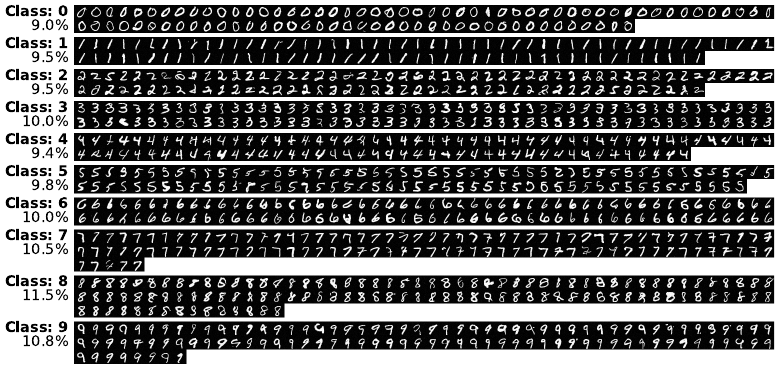}
    }\\
    \subfloat[][Adjoint matching, $a=10$, $\hat R = 0.176$]{
    \includegraphics[width=.75\linewidth]{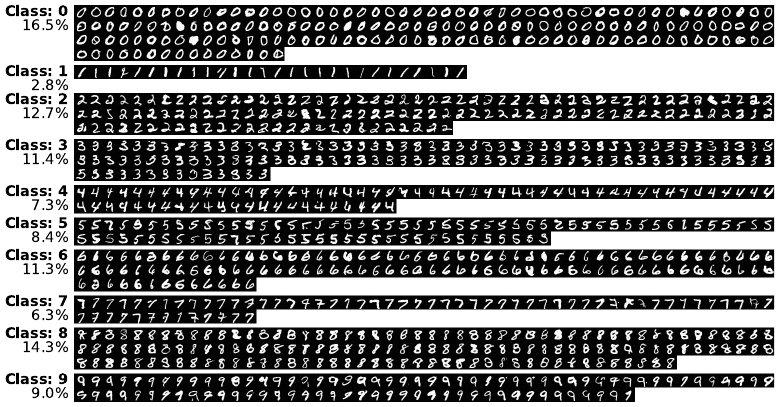}
    }\\
    \subfloat[][Weighted adjoint matching, $t_s = 0.45$, $a=227$, $\hat R = 0.180$]{
    \includegraphics[width=.75\linewidth]{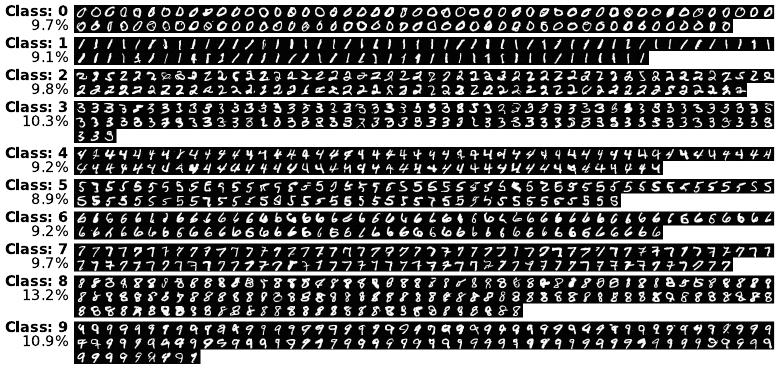}
    }
    \caption{Generated samples grouped by predicted class, for the run of each method whose average luminosity is closest to 0.18.
    Percentages are computed on the $1\,000$ displayed samples, and therefore differ from those of Table~\ref{tab:exp-MNIST_class_percentages_0.18}, computed on $50\,000$ samples.}
    \label{fig:exp-MNIST_class_balance_0.18}
\end{figure}

In Tables~\ref{tab:exp-MNIST_class_percentages_0.16}~\&~\ref{tab:exp-MNIST_class_percentages_0.18}, we have the distributions of all digits (from $50\,000$ samples classified by the pretrained classifier), for the base model, adjoint matching and weighted adjoint matching with different $t_s$. For each, we select a specific value of $a$ by looking at the one that has an achieved reward closest to $0.16$ (Table~\ref{tab:exp-MNIST_class_percentages_0.16}) and $0.18$ (Table~\ref{tab:exp-MNIST_class_percentages_0.18}).
We observed that for an average achieved reward of $0.176$, adjoint matching generates $18.0\%$ of zeros and only $3.4\%$ of ones, while weighted adjoint matching with $t_s=0.45$ reaches a reward of $0.180$ with $10.5\%$ of zeros and $9.1\%$ of ones.
Samples of the base model, adjoint matching and weighted adjoint matching with $t_s=0.45$, grouped by predicted class, are shown in Figures~\ref{fig:exp-MNIST_class_balance_0.16}~\&~\ref{fig:exp-MNIST_class_balance_0.18}.

\subsection{MNIST with darkness reward}
\label{appendix:additional_experiments:exp-MNIST-dark}
\begin{figure}
    \centering
    \includegraphics[width=0.55\linewidth]{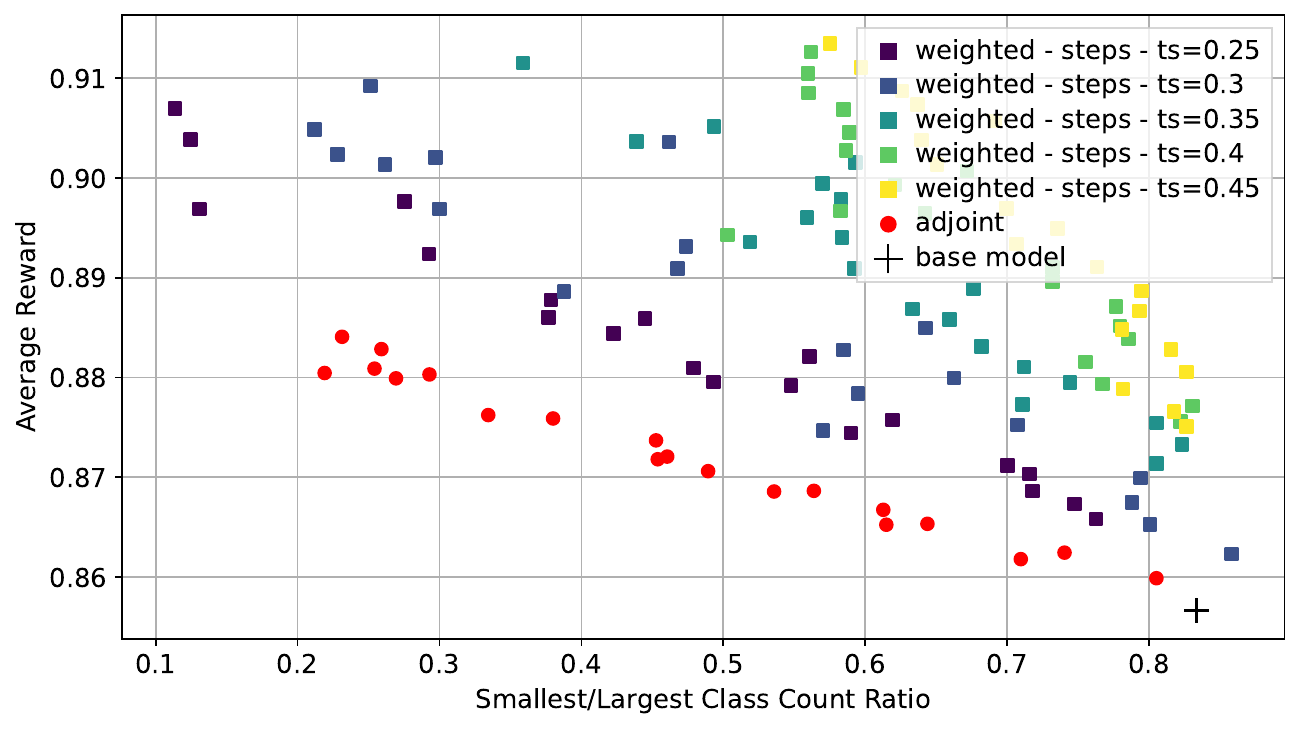}
    \caption{Average reward with respect to ratio $\frac{n_{\min}}{n_{\max}}$.}
    \label{fig:MNIST-dark:reward_by_smallest_largest_class_ratio_steps}
\end{figure}

\begin{figure}[p]
    \centering
    \captionsetup[subfigure]{justification=centering}
    \subfloat[][Base model, average reward $\hat R = 0.857$]{
    \includegraphics[width=.75\linewidth]{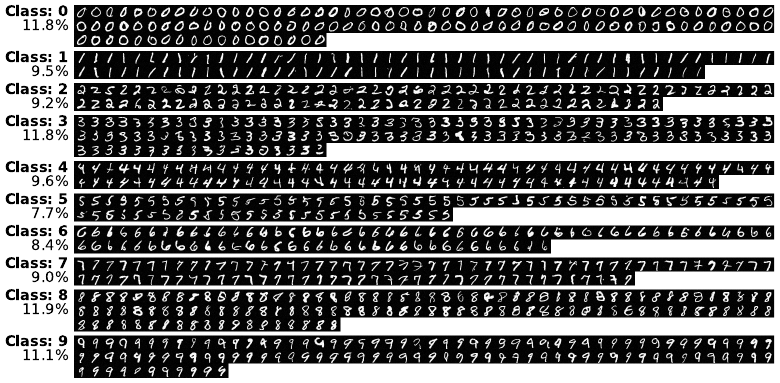}
    }\\
    \subfloat[][Adjoint matching, $a=8.6$, $\hat R = 0.880$]{
    \includegraphics[width=.75\linewidth]{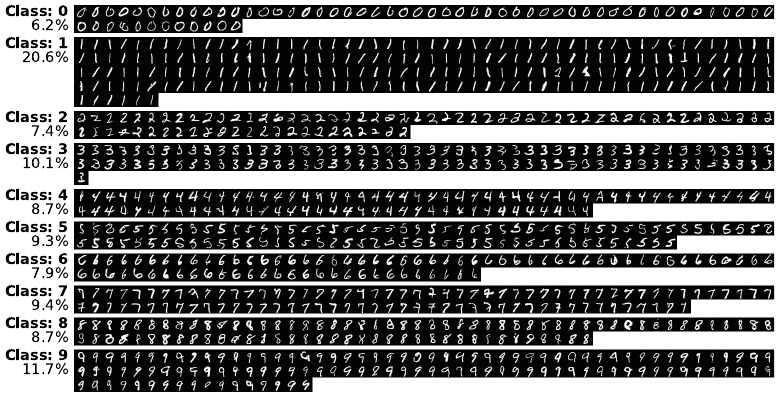}
    }\\
    \subfloat[][Weighted adjoint matching, $t_s = 0.45$, $a=160$, $\hat R = 0.881$]{
    \includegraphics[width=.75\linewidth]{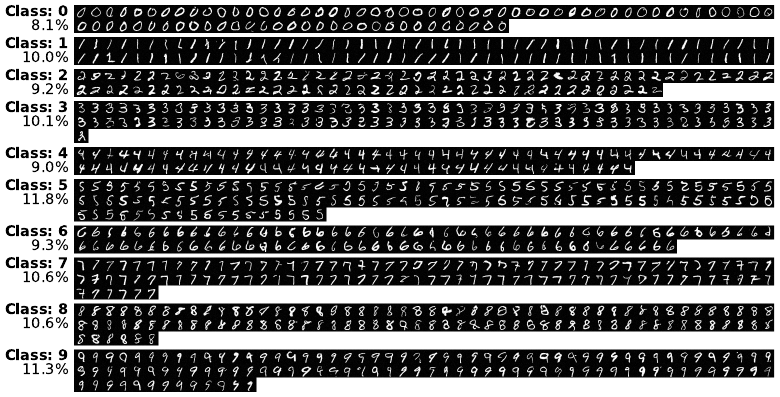}
    }
    \caption{Generated samples grouped by predicted class, for the run of each method whose average darkness is closest to 0.88.
    Percentages are computed on the $1\,000$ displayed samples, and therefore differ from those of Table~\ref{tab:exp-MNIST-dark_class_percentages_0.88}, computed on $50\,000$ samples.}
    \label{fig:exp-MNIST-dark_class_balance_0.88}
\end{figure}

We also ran an additional experiment with darkness as the reward, defined, for $x\in [0,1]^{d}$, as $r(x) = 1 - \frac{1}{d}\sum_i x_i$. Results are in Figure~\ref{fig:MNIST-dark:reward_by_smallest_largest_class_ratio_steps} and Table~\ref{tab:exp-MNIST-dark_class_percentages_0.88}.
Samples of the base model, adjoint matching and weighted adjoint matching with $t_s=0.45$, grouped by predicted class, are shown in Figure~\ref{fig:exp-MNIST-dark_class_balance_0.88}.
The conclusions are unchanged.

\begin{table}[p]
\centering
\caption{Predicted class of the generated samples, in percent, for the run of each method whose average luminosity is closest to 0.16.}
\label{tab:exp-MNIST_class_percentages_0.16}
\begin{tabular}{lrr*{10}{r}}
\toprule
 & & & \multicolumn{10}{c}{Predicted class (\%)} \\
\cmidrule(lr){4-13}
Method & $a$ & $\hat{R}$ & 0 & 1 & 2 & 3 & 4 & 5 & 6 & 7 & 8 & 9 \\
\midrule
Base & -- & 0.140 & 10.2 & 9.9 & 9.9 & 10.8 & 9.4 & 9.1 & 9.8 & 10.0 & 10.1 & 10.9 \\
\midrule
Adjoint & 5.7 & 0.159 & 14.8 & 5.8 & 12.0 & 12.0 & 7.4 & 8.5 & 10.0 & 8.0 & 12.1 & 9.4 \\
\midrule
Weighted, $t_s=0.25$ & 21 & 0.161 & 10.6 & 6.6 & 11.4 & 11.8 & 8.9 & 8.1 & 10.2 & 9.1 & 13.0 & 10.4 \\
Weighted, $t_s=0.3$ & 34 & 0.162 & 10.5 & 7.4 & 10.6 & 10.9 & 9.4 & 8.5 & 9.2 & 9.2 & 12.9 & 11.4 \\
Weighted, $t_s=0.35$ & 52 & 0.161 & 10.6 & 8.8 & 10.2 & 11.0 & 9.2 & 7.7 & 9.8 & 9.0 & 12.3 & 11.2 \\
Weighted, $t_s=0.4$ & 80 & 0.160 & 10.5 & 9.5 & 10.0 & 10.7 & 9.2 & 8.4 & 9.7 & 9.3 & 11.5 & 11.2 \\
Weighted, $t_s=0.45$ & 120 & 0.160 & 10.3 & 9.7 & 9.8 & 10.6 & 9.1 & 8.5 & 9.8 & 9.6 & 11.2 & 11.3 \\
\bottomrule
\end{tabular}
\end{table}

\begin{table}[p]
\centering
\caption{Predicted class of the generated samples, in percent, for the run of each method whose average luminosity is closest to 0.18.}
\label{tab:exp-MNIST_class_percentages_0.18}
\begin{tabular}{lrr*{10}{r}}
\toprule
 & & & \multicolumn{10}{c}{Predicted class (\%)} \\
\cmidrule(lr){4-13}
Method & $a$ & $\hat{R}$ & 0 & 1 & 2 & 3 & 4 & 5 & 6 & 7 & 8 & 9 \\
\midrule
Base & -- & 0.140 & 10.2 & 9.9 & 9.9 & 10.8 & 9.4 & 9.1 & 9.8 & 10.0 & 10.1 & 10.9 \\
\midrule
Adjoint & 10 & 0.176 & 18.0 & 3.4 & 12.5 & 12.3 & 7.0 & 7.9 & 10.7 & 6.5 & 13.1 & 8.6 \\
\midrule
Weighted, $t_s=0.25$ & 37 & 0.181 & 12.0 & 3.4 & 13.7 & 11.3 & 8.2 & 7.8 & 8.0 & 7.7 & 17.4 & 10.6 \\
Weighted, $t_s=0.3$ & 57 & 0.180 & 11.9 & 6.1 & 11.2 & 11.6 & 8.1 & 6.4 & 8.9 & 8.4 & 16.0 & 11.2 \\
Weighted, $t_s=0.35$ & 92 & 0.180 & 10.6 & 7.8 & 11.0 & 10.7 & 8.5 & 7.6 & 9.7 & 8.6 & 14.2 & 11.4 \\
Weighted, $t_s=0.4$ & 143 & 0.180 & 10.7 & 8.4 & 10.4 & 10.5 & 8.6 & 7.6 & 9.8 & 9.1 & 13.3 & 11.6 \\
Weighted, $t_s=0.45$ & 227 & 0.180 & 10.5 & 9.1 & 10.2 & 10.3 & 8.9 & 7.9 & 9.7 & 9.3 & 12.4 & 11.5 \\
\bottomrule
\end{tabular}
\end{table}

\begin{table}[p]
\centering
\caption{Predicted class of the generated samples, in percent, for the run of each method whose average darkness is closest to 0.88.}
\label{tab:exp-MNIST-dark_class_percentages_0.88}
\begin{tabular}{lrr*{10}{r}}
\toprule
 & & & \multicolumn{10}{c}{Predicted class (\%)} \\
\cmidrule(lr){4-13}
Method & $a$ & $\hat{R}$ & 0 & 1 & 2 & 3 & 4 & 5 & 6 & 7 & 8 & 9 \\
\midrule
Base & -- & 0.857 & 10.4 & 10.1 & 9.4 & 10.8 & 9.5 & 9.0 & 9.9 & 9.8 & 10.3 & 10.7 \\
\midrule
Adjoint & 8.6 & 0.880 & 5.3 & 19.6 & 7.8 & 9.9 & 9.6 & 9.6 & 8.9 & 10.8 & 7.8 & 10.6 \\
\midrule
Weighted, $t_s=0.25$ & 25 & 0.880 & 8.8 & 14.8 & 8.1 & 10.4 & 12.0 & 9.0 & 7.3 & 10.4 & 7.6 & 11.6 \\
Weighted, $t_s=0.3$ & 43 & 0.880 & 9.3 & 12.2 & 8.8 & 11.1 & 10.2 & 10.0 & 9.6 & 10.7 & 8.1 & 10.1 \\
Weighted, $t_s=0.35$ & 63 & 0.880 & 9.8 & 11.2 & 9.0 & 10.5 & 10.6 & 10.5 & 9.4 & 10.4 & 8.4 & 10.2 \\
Weighted, $t_s=0.4$ & 98 & 0.879 & 10.1 & 10.6 & 9.4 & 11.0 & 8.5 & 9.8 & 10.0 & 10.4 & 9.1 & 11.1 \\
Weighted, $t_s=0.45$ & 160 & 0.881 & 10.1 & 10.3 & 9.8 & 11.0 & 9.9 & 9.9 & 9.1 & 10.3 & 9.2 & 10.5 \\
\bottomrule
\end{tabular}
\end{table}

\end{document}